%% file: main.tex
\documentclass[10pt,twocolumn,letterpaper]{article}

\usepackage[pagenumbers]{wacv}

\input{preamble}
\usepackage[T1,OT1]{fontenc} 
\DeclareTextCommandDefault{\k}[1]{{\fontencoding{T1}\selectfont\k{#1}}}

\definecolor{wacvblue}{rgb}{0.21,0.49,0.74}
\usepackage[pagebackref,breaklinks,colorlinks,allcolors=wacvblue]{hyperref}
\usepackage{multirow}
\usepackage{pgfplots}
\usepackage{graphicx}
\usepackage[export]{adjustbox}
\usepackage{bm}
\usepackage{caption}
\usepackage{rotating}
\usepackage{array}
\usepackage{xcolor}
\usepackage{float}

\newlength{\imgw}
\newlength{\imgh}
\definecolor{correct}{rgb}{0.00,0.60,0.00}   %
\definecolor{wrong}{rgb}{0.85,0.00,0.00}     %
\definecolor{querycol}{rgb}{0.10,0.30,0.80}  %
\definecolor{milestone}{rgb}{0.80,0.40,0.00} %

\def\wacvPaperID{296}
\def\confName{WACV}
\def\confYear{2027}

\title{Learning Discriminative Prototypes via Boundary Repulsion MixUp for Object Re-Identification}

\author{Athena Psalta \qquad Vasileios Tsironis \qquad Konstantinos Karantzalos\\
National Technical University of Athens\\
Iroon Polytechneiou 9, Athens 15780, Greece\\
{\tt\small \{psaltaath, tsironisbi, karank\}@central.ntua.gr}
}

\begin{document}
\maketitle
\input{sec/0_abstract}
\input{sec/1_intro}

\input{sec/2_relatedwork}
\input{sec/3_methodology_revised}
\input{sec/4_experiments}

\input{sec/5_conclusions}

{
    \small
    \bibliographystyle{ieeenat_fullname}
    \bibliography{main}
}

\clearpage
\appendix

\setcounter{topnumber}{4}
\setcounter{bottomnumber}{4}
\setcounter{totalnumber}{6}
\renewcommand{\topfraction}{0.95}
\renewcommand{\bottomfraction}{0.95}
\renewcommand{\textfraction}{0.05}
\renewcommand{\floatpagefraction}{0.85}
\renewcommand{\dbltopfraction}{0.95}
\renewcommand{\dblfloatpagefraction}{0.5}
\makeatletter\setlength{\@dblfptop}{0pt}\makeatother %

\input{sec/00_supplementary}

\input{supp/veri_retrieval_comparison}

\input{supp/msmt17_retrieval_comparison}

\end{document}

%% file: sec/0_abstract.tex
\begin{abstract}

Object re-identification requires embeddings that are robust to intra-class variation in viewpoint, illumination, and occlusion, yet preserve sharp separation between visually similar identities. The hardest errors arise precisely where confusable identities meet, a region that pairwise losses shape only implicitly. We present PBR-ReID, a prototype-based framework that makes this inter-class geometry an explicit learning target. Each identity is represented by a learnable prototype jointly optimized with a DiNOv3 backbone and the resulting prototype similarity matrix reveals, at every training step, which identities are most confusable. On this signal we build Boundary Repulsion MixUp (BRM), a geometry-aware regularizer that converts standard MixUp's weakness into a discriminative signal. Rather than interpolating samples across confusable identities, which blurs the boundaries that matter most, BRM constructs an explicit boundary proxy between each confusable pair and repels their features away from it, converting boundary ambiguity into a discriminative signal. We adopt a two-stage training protocol consisting of a prototype warmup phase followed by EMA-stabilized joint fine-tuning. Extensive experiments on person and vehicle ReID benchmarks demonstrate strong performance, with thorough ablations confirming each component's contribution. Code is available at \url{https://github.com/psaltaath/pbr-reid}.

\end{abstract}

%% file: sec/1_intro.tex
\section{Introduction}
\label{sec:intro}

Object Re-Identification (ReID) is an important component of intelligent surveillance \cite{pandeeswari2025deep}, autonomous driving \cite{li2025clip} and multi-camera tracking systems \cite{maggiolino2023deep}, where fine-grained identity matching across non-overlapping cameras must remain robust to viewpoint, illumination and occlusion while precisely separating visually similar identities.
Recent advances in Vision Transformers \cite{He_2021_ICCV, wang2024drformer}, CLIP-driven alignment \cite{li2022clip, climb} and part-based attention \cite{AAformer} have pushed retrieval accuracy to impressive levels, yet a more targeted question remains underexplored: \textit{should identity representatives be post-hoc feature aggregates, or learnable parameters that actively structure the embedding space?}

The most straightforward representation treats each gallery image as an independent representative which is maximally expressive, but outlier-sensitive and linear in gallery size. Aggregating per-class embeddings into a single centroid proves surprisingly effective since it improves both accuracy and outlier robustness while reducing gallery search complexity \cite{wieczorek2021unreasonable}, yet centroids are static post-hoc aggregates. Thus, they are confined to the convex hull of observed features and brittle under multi-modal intra-class distributions. Prototype-based representations have been explored across ReID as memory anchors for unsupervised and continual learning \cite{xu2024distribution, dai2022cluster}, as prototypical contrastive targets \cite{li2023prototypical}, and as learnable classifier weights retained at inference \cite{wang2023rethinking}. Generalized prototypes constructed from gallery features at inference push past the centroid ceiling \cite{muzaddid2025person}, but tie the model to the gallery at test time and leave training-time geometry uncontrolled. Across all these cases, however, prototypes remain reactive aggregates or inference-time constructs, decoupled from the inter-class geometry they induce and never used as a discriminative signal during training.

These failures are not spread uniformly across the embedding space but concentrate at the boundaries between confusable classes. Classification errors under distribution shift cluster near semantically similar classes and models with larger inter-class margins are substantially more robust to such shifts \cite{yang2025gain}. The analogue in ReID is direct, since the most damaging retrieval failures occur between visually confusable identities (e.g. persons in matching attire, vehicles of identical design and color) whose embeddings lie near shared decision boundaries. Standard regularization techniques can deepen this fragility rather than relieve it. Interpolation-based methods such as MixUp \cite{zhang2017mixup} are agnostic to class geometry, sampling mixed features uniformly between pairs, including between highly confusable identities, and thus placing synthetic samples in the most ambiguous regions, where they reinforce rather than resolve boundary confusion. The common thread is that without a stable, explicit representation of class structure during training, the boundaries between genuinely confusable identities are never modeled, and the margins they demand are never enforced.

In this work, we introduce \textbf{Prototype-based Boundary Repulsion for ReID (PBR-ReID)}, a framework built on two components that treats identity representatives as learnable parameters actively structuring the embedding space, making the inter-class geometry itself the learning target. First, we attach to each identity $k$ a \textit{learnable prototype} $\mathbf{p}_k \in \mathbb{R}^d$, a free parameter optimized jointly with a DINOv3 \cite{simeoni2025dinov3} backbone that, unlike a post-hoc centroid, adapts to the global inter-class geometry and makes pairwise identity relationships explicit through the prototype similarity matrix $\mathbf{S}_{ij} = \hat{\mathbf{p}}_i^\top \hat{\mathbf{p}}_j$. The prototypes are purely a training-time mechanism, shaping the feature geometry during optimization and discarded at inference, where retrieval relies on backbone features alone. Second, we introduce \textit{Boundary Repulsion MixUp} (BRM), which reads $\mathbf{S}$ to single out the most confusable identity pairs, interpolates boundary proxies between their prototypes and penalizes any feature that encroaches upon these proxies, converting boundary ambiguity into an explicit discriminative signal. We train PBR-ReID with a two-stage scheme of prototype warmup followed by EMA-stabilized joint fine-tuning and report strong performance across five person ReID benchmarks (Market-1501 \cite{market1501}, DukeMTMC-reID \cite{dukemtmc}, MSMT17 \cite{msmt17}, CUHK03 \cite{cuhk03}, IUST \cite{iust}) and two vehicle ReID benchmarks (VeRi-776 \cite{veri776}, VehicleID \cite{vehicleid}), with ablations isolating the contribution of each component.

%% file: sec/2_relatedwork.tex
\section{Related Work}
\label{sec:relatedwork}

\noindent \textbf{ReID \& Identity Representations.}
ViT \cite{dosovitskiy2020image} backbones are now standard for ReID \cite{He_2021_ICCV}, with subsequent work diversifying along CLIP-driven vision-language alignment \cite{radford2021learning, kim2026dico, li2022clip} for semantic supervision without explicit part annotations, self-supervised or prompt-driven pre-training \cite{chen2023beyond, zhizhong2025vicp} for cross-domain generalization and attention- and part-based decomposition \cite{AAformer, bansal2022cloth} for fine-grained local discrimination. Across these directions the backbone is treated as the primary driver of discriminability, leaving class structure to emerge implicitly from pairwise losses. We build on DINOv3 \cite{simeoni2025dinov3}, a self-supervised ViT whose distillation pre-training yields semantically rich representations well-suited to fine-grained retrieval, and pair it with prototypes that make class structure explicit. 

As class-level anchors, prototypes enforce intra-class compactness and inter-class separation \cite{snell2017prototypical, jin2023prototypical} and a central design choice is whether they are estimated dynamically from features or fixed as pre-specified targets, the latter avoiding the estimation bias of feature-derived centroids under long-tailed distributions \cite{ge2024beyond}. The most standard approach optimizes the prototype matrix as the weights of a margin-based softmax classifier \cite{deng2019arcface, wang2018cosface, deng2021variational} or pair-similarity objective \cite{sun2020circle}, imposing an angular or cosine separation fixed a priori rather than one informed by which identities are actually confusable. Recently, ReID methods have repurposed prototypes as memory anchors for unsupervised and continual learning \cite{xu2024distribution, dai2022cluster, li2021prototypical} or as learnable classifier weights retained at inference \cite{wang2023rethinking, muzaddid2025person}. While these methods establish prototypes as effective class representatives, the prototype set is consumed downstream rather than optimized so that its own pairwise structure becomes the supervisory signal. Moreover, methods that reuse prototypes at test time rely on them at inference to compute similarity features; our use of prototypes is orthogonal, since they shape the embedding geometry only on training and are discarded at inference. We treat that geometry as the learning target, optimizing the prototype configuration so that its mutual structure becomes an explicit source of discrimination.

\noindent \textbf{Regularization in ReID.}
Metric learning objectives such as triplet loss \cite{schroff2015facenet} with its hard-mining variants \cite{zhao2020deep} and center loss \cite{wen2019comprehensive}, together with their ReID adaptations \cite{Yuan_2020_CVPR_Workshops, hermans2017defense}, enforce intra-class compactness and inter-class separation, complemented by geometry-aware regularizers such as hypersphere projections \cite{fan2019spherereid}, momentum-based feature consistency \cite{11036118} and structure-aware triplet mining \cite{ghosh2023relation}. Augmentation-based regularization, including patch-wise high-frequency augmentation \cite{zhang2023pha}, intra/cross-modality consistency \cite{tan2024rle, chen2025consistency} and MixUp \cite{zhang2017mixup} with its variants \cite{pinto2022using, venkataramanan2021takes}, has inspired ReID methods that bridge intra/inter-identity gaps, with recent extensions such as part descriptor mixing \cite{kim2023partmix}, centroid-guided sampling \cite{mamedov2025remix} and dynamic mixing \cite{mamedov2025dynamix} addressing domain challenges. However, these objectives either operate on sampled triplets or interpolate pairs uniformly, leaving confusion between geometrically confusable identities near embedding boundaries unaddressed. BRM instead reads the global inter-class structure from the prototype similarity matrix and targets the most ambiguous pairs directly, repelling features from boundary proxies interpolated between confusable prototypes and converting boundary ambiguity into an explicit discriminative signal.

%% file: sec/3_methodology_revised.tex
\section{Methodology}
\label{sec:methodology}

\begin{figure*}[t]
  \centering
  \includegraphics[width=0.60\textwidth]{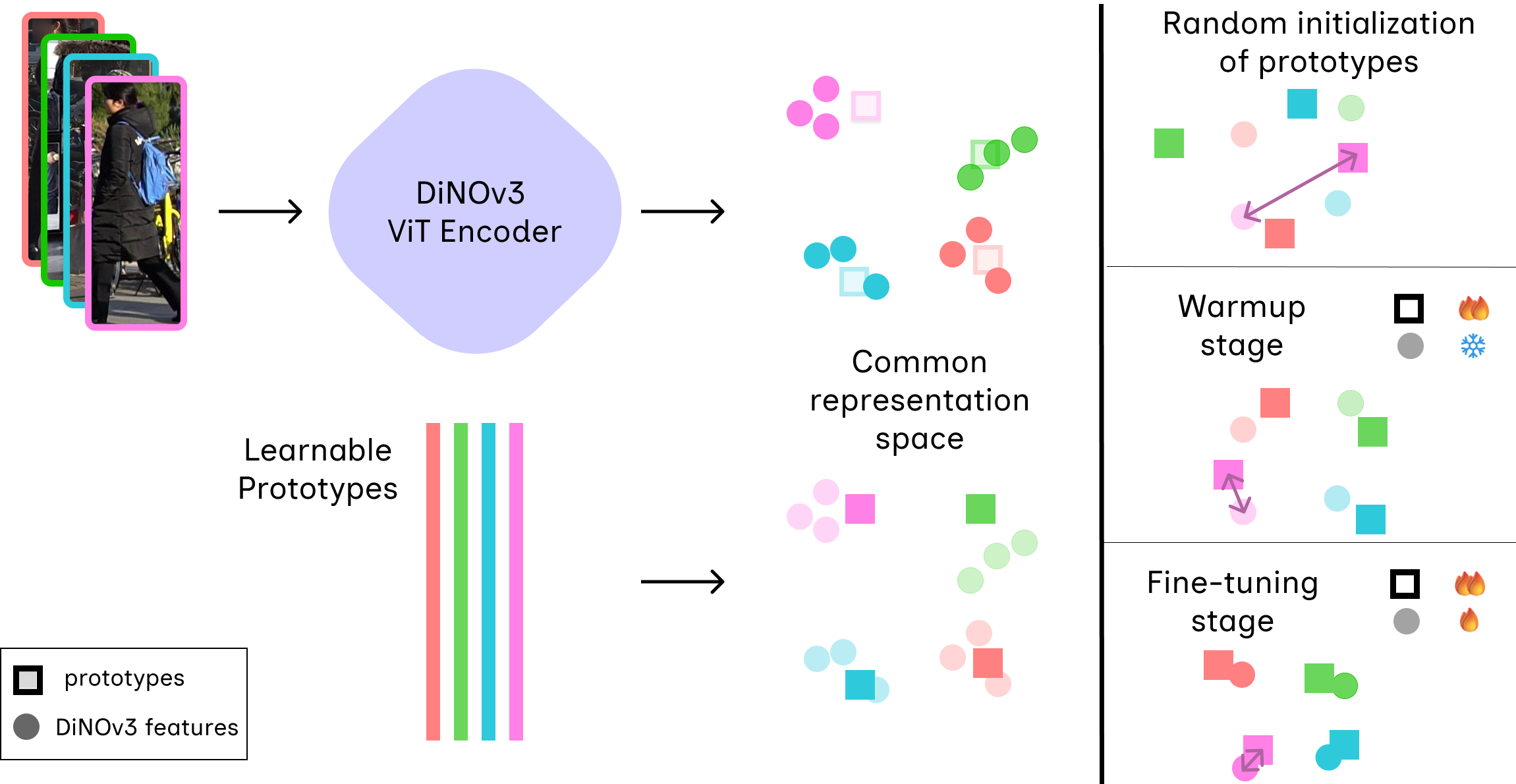}
  \caption{\textbf{PBR-ReID Framework.} Input images are passed through a DiNOv3 ViT backbone to produce normalized embeddings, jointly optimized with a learnable prototype matrix (one vector per identity) that anchors each class in the embedding space. The staged training progression (right) illustrates how embedding geometry evolves from random initialization through prototype warmup to joint fine-tuning, 
  with features forming compact intra-class clusters separated by enforced inter-class margins. Prototypes act only during training and are discarded at inference.}
  \label{fig:methodology1}
\end{figure*}

\subsection{Feature Representation}
\label{subsec:representation}

\noindent\textbf{Feature Extraction.}
We employ DiNOv3~\cite{simeoni2025dinov3} as our backbone encoder. For an input image $\mathbf{x} \in \mathbb{R}^{H \times W \times 3}$, the backbone produces a global embedding $\mathbf{f} = \phi_{\text{DINOv3}}(\mathbf{x}) \in \mathbb{R}^{d}$. All embeddings are $L_2$ unit-normalized.

\noindent\textbf{Prototype-Based Representation.}
Each identity $k \in \{1, \ldots, K\}$ is represented by a learnable
prototype vector $\mathbf{p}_k \in \mathbb{R}^d$, forming the prototype matrix
$\mathbf{P} \in \mathbb{R}^{K \times d}$. Prototypes are normalized to $\mathbb{S}^{d-1}$ yielding $\hat{\mathbf{p}}_k$.
Let $\mathcal{K}_B$ denote the set of
identities present in the current mini-batch, with $|\mathcal{K}_B| = K'$. The
pairwise inter-prototype similarities among the batch identities form the
prototype similarity matrix $\mathbf{S} \in \mathbb{R}^{K' \times K'}$, with
$\mathbf{S}_{ij} = \hat{\mathbf{p}}_i^\top \hat{\mathbf{p}}_j$ for
$i \neq j$ and $i,j \in \mathcal{K}_B$, encoding the inter-class
confusion structure among the sampled identities and recomputed at every
training step as $\mathbf{S} = \hat{\mathbf{P}}_B\hat{\mathbf{P}}_B^\top$,
where $\hat{\mathbf{P}}_B \in \mathbb{R}^{K' \times d}$ stacks the row-normalized
prototypes of the batch identities. Restricting $\mathbf{S}$ to the $K'$ sampled
identities keeps its construction at $\mathcal{O}(K'^2 d)$ cost per step,
independent of the total number of identities $K$, so all $\mathbf{S}$-derived
quantities scale to datasets with tens of thousands of identities without
materializing a global $K \times K$ matrix. To stabilize the representation
space, we isolate the prototype updates from the rest of the network through a
dedicated optimizer and a separate learning rate schedule.

\subsection{Learning Framework}
\label{subsec:framework}

\subsubsection{Training Objective}
\label{subsubsec:objective}

We optimize the encoder weights and the prototypes jointly, under independent training dynamics for stable convergence. The full training objective is
\begin{equation}
\label{eq:total-loss}
\begin{aligned}
    \mathcal{L} = {}& \lambda_{\mathrm{bht}}\,\mathcal{L}_{\mathrm{bht}}
                + \lambda_{\mathrm{fpt}}\,\mathcal{L}_{\mathrm{fpt}}
               + \lambda_{\mathrm{cen}}\,\mathcal{L}_{\mathrm{cen}} \\
               & + \lambda_{\mathrm{norm}}\,\mathcal{L}_{\mathrm{norm}}
               + \lambda_{\mathrm{BRM}}\,\mathcal{L}_{\mathrm{BRM}},
\end{aligned}
\end{equation}
a weighted sum of five terms with scalar weights $\lambda_{\bullet}$: a batch-hard feature triplet $\mathcal{L}_{\text{bht}}$, a prototype-anchored triplet $\mathcal{L}_{\text{fpt}}$, a prototype center loss $\mathcal{L}_{\text{cen}}$, a soft norm regularizer $\mathcal{L}_{\text{norm}}$, and the Boundary Repulsion MixUp regularizer $\mathcal{L}_{\text{BRM}}$ that vacates the confusion zone between visually similar identities (Sec.~\ref{subsubsec:brm}). The center loss and the positive triplet terms enforce intra-class \textit{compactness}, while their hard-negative terms and the BRM regularizer enforce inter-class \textit{separation}. These two are coupled, since compactness without a separation reference collapses the embedding, whereas separation beyond the reachable intra-class spread distorts it. The terms are activated in stages, with $\mathcal{L}_{\text{bht}}$ and $\mathcal{L}_{\text{BRM}}$ enabled only during joint fine-tuning (Sec.~\ref{subsubsec:training}).

Supervision is provided by a class-balanced sampling strategy. Each mini-batch draws $K'$ identities with $N_k$ instances each ($B = K' \times N_k$), guaranteeing valid positive pairs for the center loss, negative diversity for hard-negative mining (HNM), and a stable online estimate of the batch prototype similarity matrix $\mathbf{S}$. We detail each term below.

\noindent \textbf{Batch-Hard Feature Triplet.}
To tighten the residual intra-class spread we add an instance-anchored batch-hard triplet~\cite{hermans2017defense} computed directly between features. For each anchor $\mathbf{f}_i$ we mine its hardest positive and hardest negative among the remaining batch features,
\begin{equation}
\begin{aligned}
    j_i^{+} = \arg\max_{j\,:\,y_j = y_i,\, j \neq i} d(\mathbf{f}_i, \mathbf{f}_j), \\
    j_i^{-} = \arg\min_{j\,:\,y_j \neq y_i} d(\mathbf{f}_i, \mathbf{f}_j),
\end{aligned}
\end{equation}
and penalize the margin violation between them,
\begin{equation}
\label{eq:bht-loss}
    \mathcal{L}_{\mathrm{bht}} = \frac{1}{|\mathcal{A}|} \sum_{i \in \mathcal{A}}
        \left[ d(\mathbf{f}_i, \mathbf{f}_{j_i^{+}})
             - d(\mathbf{f}_i, \mathbf{f}_{j_i^{-}})
             + m_{\mathrm{bht}} \right]_+,
\end{equation}
where $\mathcal{A}$ is the set of anchors with at least one positive and one negative in the batch, and $m_{\mathrm{bht}}$ is the margin. Its hardest positive is typically the cross-camera instance, so the term compresses cross-camera spread; being purely instance-to-instance, it is active only during joint fine-tuning.

\noindent \textbf{Feature-Prototype Triplet Loss.}
We employ a triplet loss in a \emph{feature-prototype} mining mode: each class prototype is the anchor, and the positive and negative are mined from the batch features in a batch-hard fashion. For each identity $k \in \mathcal{K}_B$, the anchor is its prototype $\mathbf{p}_k$, the hardest positive is the most distant same-class feature, and the hardest negative is the closest different-class feature:
\begin{equation}
\begin{aligned}
    i_k^{+} &= \arg\max_{i\,:\,y_i = k} d(\mathbf{p}_k, \mathbf{f}_i), \\
    i_k^{-} &= \arg\min_{j\,:\,y_j \neq k} d(\mathbf{p}_k, \mathbf{f}_j),
\end{aligned}
\end{equation}
where $d(\cdot, \cdot) = 1 - \cos(\cdot, \cdot)$ is the cosine distance. The loss enforces a margin $m_{\text{tri}}$, averaged over the batch identities and scaled by an adaptive weight $w_k$ that focuses learning on the most confusable instances:
\begin{equation}
\begin{aligned}
    \mathcal{L}_{\text{fpt}} &= \frac{1}{K'}\sum_{k=1}^{K'} w_k \left[ d(\mathbf{p}_k, \mathbf{f}_{i_k^{+}}) - d(\mathbf{p}_k, \mathbf{f}_{i_k^{-}}) + m_{\text{tri}} \right]_+, \\
    w_k &= \exp\!\left(\gamma_{\text{triplet}} \cdot \mathbf{p}_k^\top \mathbf{f}_{i_k^{-}}\right),
\end{aligned}
\end{equation}
where $[\cdot]_+ = \max(0, \cdot)$, $\gamma_{\text{triplet}}$ controls the weighting strength ($w_k = 1$ disables it), and the negative is an instance-level hard sample from the current batch, up-weighting anchors whose nearest intruding identity is most confusable.

\noindent \textbf{Prototype Center Loss.}
To complement separation, we penalize intra-class dispersion by pulling features toward their class prototype. For identity $k$ with index set $\mathcal{B}_k = \{i : y_i = k\}$, we summarize the class by the intrinsic (Fr\'echet) mean of its features, which stays on $\mathbb{S}^{d-1}$ whereas the Euclidean mean does not:
\begin{equation}
    \boldsymbol{\mu}_k = \arg\min_{\boldsymbol{\mu} \in \mathbb{S}^{d-1}}
        \sum_{i \in \mathcal{B}_k} \arccos(\boldsymbol{\mu}^\top \mathbf{f}_i)^2.
\end{equation}
The per-class loss is the cosine distance $\ell_k = 1 - \mathbf{p}_k^\top \boldsymbol{\mu}_k$, re-weighted by two detached factors: a \emph{similarity} weight $w_k^{\mathrm{sim}} = \exp(\gamma_{\text{center}}^{\text{sim}}\,\bar{s}_k)$ that emphasizes identities with high average inter-prototype similarity $\bar{s}_k = \tfrac{1}{K'-1}\sum_{k'\in\mathcal{K}_B,\,k'\neq k}\mathbf{S}_{k,k'}$, and a \emph{variance} weight $w_k^{\mathrm{var}} = 1 + \gamma_{\text{center}}^{\text{var}}\,\sigma_k^2$ that grows with the intra-class dispersion $\sigma_k^2 = \tfrac{1}{|\mathcal{B}_k|}\sum_{i\in\mathcal{B}_k}\|\mathbf{f}_i - \boldsymbol{\mu}_k\|_2^2$. The loss aggregates over the $K'$ batch identities:
\begin{equation}
    \mathcal{L}_{\mathrm{cen}} = \frac{1}{K'}
        \sum_{k=1}^{K'} w_k^{\mathrm{sim}} \cdot w_k^{\mathrm{var}}
        \cdot \left(1 - \mathbf{p}_k^\top \boldsymbol{\mu}_k\right),
\end{equation}
with $w_k^{\mathrm{sim}} = w_k^{\mathrm{var}} = 1$ when the respective weighting terms are disabled.

\noindent \textbf{Feature Normalization Loss.}
We add a soft unit-norm regularizer on both features and prototypes, penalizing the gap between each vector and its $\ell_2$-normalized version:
\begin{equation}
    \mathcal{L}_{\text{norm}} =
        \frac{1}{B} \sum_{i=1}^{B} \|\mathbf{f}_i - \hat{\mathbf{f}}_i\|_2
        + \frac{1}{K} \sum_{k=1}^{K} \|\mathbf{p}_k - \hat{\mathbf{p}}_k\|_2 ,
\end{equation}
which keeps the backbone and prototypes near the sphere, complementing explicit normalization.

\begin{figure*}[t]
  \centering
  \includegraphics[width=0.8\textwidth]{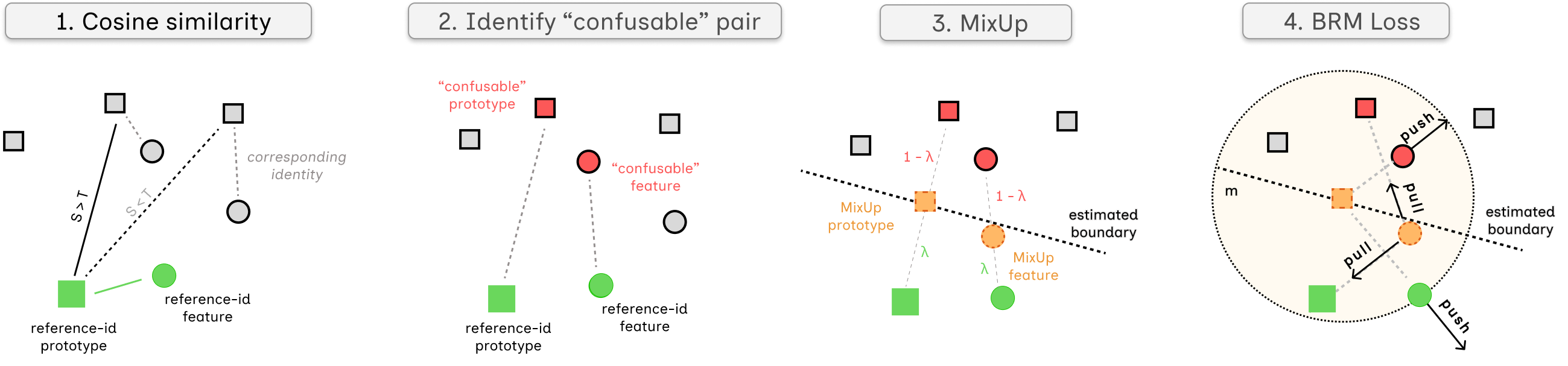}
  \caption{\textbf{Boundary Repulsion MixUp (BRM).} BRM operates in four stages. First, pairwise prototype cosine similarities are computed for all identities in the batch. Second, confusable pairs (those exceeding similarity threshold $\tau$) are identified. Third, for each confusable pair, a boundary-proxy prototype is constructed by linearly interpolating the two class prototypes with a Beta-sampled coefficient $\lambda$, which also mixes the corresponding features. Fourth, the BRM loss combines hard-target attraction terms, which pull the mixed feature toward both participating prototypes, with repulsion terms, which penalize original features lying within margin $m_{\mathrm{BRM}}$ of the boundary-proxy prototype.}
  \label{fig:methodology2}
\end{figure*}

\subsubsection{Boundary Repulsion MixUp (BRM)}
\label{subsubsec:brm}
BRM handles systematic confusions between visually similar identities. Unlike standard MixUp, which can reinforce ambiguity by synthesizing samples inside the confusion zone between similar identities, BRM constructs an explicit proxy for the decision boundary between confusable pairs and repels original samples away from it, while retaining MixUp's manifold regularization through hard-target attraction (Fig.~\ref{fig:methodology2}).

For each batch, BRM mines the confusable pairs over the batch identities,
\begin{equation}
    \mathcal{P} = \bigl\{(i,j) : i < j,\ y_i \neq y_j,\
        \mathbf{S}_{y_i,\, y_j} > \tau \bigr\},
\end{equation}
where $\tau \in [0,1]$ is a similarity threshold; when $\mathcal{P} = \emptyset$, BRM falls back to standard random MixUp so regularization is never absent. For each pair, a coefficient $\lambda_{ij} \sim \mathrm{Beta}(\alpha, \alpha)$ mixes the features and, with the same coefficient, the participating prototypes into a \emph{boundary-proxy prototype}:
\begin{equation}
\label{eq:boundary-proto}
\begin{aligned}
    \tilde{\mathbf{f}}_{ij} &=
        \lambda_{ij}\,\mathbf{f}_i + (1 - \lambda_{ij})\,\mathbf{f}_j, \\
    \mathbf{b}_{ij} &=
        \lambda_{ij}\,\mathbf{p}_{y_i} + (1 - \lambda_{ij})\,\mathbf{p}_{y_j}.
\end{aligned}
\end{equation}
The proxy $\mathbf{b}_{ij}$ approximates the interpolated decision boundary between $y_i$ and $y_j$; a model mapping a sample near it is maximally uncertain between the two identities, so the repulsion makes this region forbidden.

The per-pair loss combines attraction and repulsion. Two \textbf{attraction terms} implement hard-target MixUp, pulling $\tilde{\mathbf{f}}_{ij}$ toward both prototypes,
\begin{equation}
    \ell_{ij}^{\mathrm{att}} =
        \lambda_{ij}\,\mathcal{L}\!\left(\tilde{\mathbf{f}}_{ij},\,\mathbf{p}_{y_i}\right)
        + (1-\lambda_{ij})\,\mathcal{L}\!\left(\tilde{\mathbf{f}}_{ij},\,\mathbf{p}_{y_j}\right),
\end{equation}
with $\mathcal{L}(\cdot,\cdot) = 1 - \cos(\cdot,\cdot)$, preserving MixUp's smoothing of the manifold. Two \textbf{repulsion terms} push the original features off the proxy via a hinge with margin $m_{\text{BRM}} > 0$,
\begin{equation}
\begin{aligned}
    \ell_{ij}^{\mathrm{rep}} = {}&
        \max\!\left(0,\, m_{\text{BRM}} - \|\mathbf{f}_i - \mathbf{b}_{ij}\|_2\right) \\
        & + \max\!\left(0,\, m_{\text{BRM}} - \|\mathbf{f}_j - \mathbf{b}_{ij}\|_2\right),
\end{aligned}
\end{equation}
enforcing a buffer zone around the confusion region. A fifth contribution is implicit: since $\mathbf{b}_{ij}$ is differentiable in both prototypes (Eq.~\ref{eq:boundary-proto}), the repulsion gradient also drives $\mathbf{p}_{y_i}$ and $\mathbf{p}_{y_j}$ apart.

We add an explicit prototype separation with a data-adaptive margin. Pushing two targets farther apart than the intra-class spread they must clear forces the encoder to chase directions the data manifold cannot supply, distorting the embedding; separation should therefore stop once it clears that spread. Following the robust-ranking intuition $1-\cos(\mathbf{p}_a,\mathbf{p}_b) \gtrsim \hat{\sigma}_a + \hat{\sigma}_b$, we separate each confusable class pair until it clears $m^{\mathrm{sep}}_{ab} = \gamma_{\mathrm{sep}}(\hat{\sigma}_a + \hat{\sigma}_b)$ ($\gamma_{\mathrm{sep}} \geq 1$), where $\hat{\sigma}_k$ is a detached EMA of the intra-class angular spread $\tfrac{1}{|\mathcal{B}_k|}\sum_{i\in\mathcal{B}_k}(1-\cos(\mathbf{f}_i,\boldsymbol{\mu}_k))$. Gradients flow only into the prototypes, so confusable targets separate just to the margin rather than maximally, keeping them reachable.

The full BRM objective aggregates the per-pair attraction and repulsion over $\mathcal{P}$ with the adaptive-margin separation over the confusable class pairs $\mathcal{P}_S = \{(a,b) : a < b,\ a,b \in \mathcal{K}_B,\ \mathbf{S}_{ab} > \tau\}$ drawn from the same batch identities:
\begin{equation}
\begin{aligned}
    \mathcal{L}_{\mathrm{BRM}} = {}& \frac{1}{|\mathcal{P}|} \sum_{(i,j)\in\mathcal{P}}
        \bigl( \ell_{ij}^{\mathrm{att}} + \beta\,\ell_{ij}^{\mathrm{rep}} \bigr) \\
    & + \frac{1}{|\mathcal{P}_S|} \sum_{(a,b)\in\mathcal{P}_S}
        \bigl[\, m^{\mathrm{sep}}_{ab} - \bigl(1 - \cos(\mathbf{p}_a,\mathbf{p}_b)\bigr) \,\bigr]_+ ,
\end{aligned}
\end{equation}
where $\beta > 0$ trades off MixUp regularization against boundary hardening, and BRM enters the full objective (Eq.~\ref{eq:total-loss}) through the single scalar weight $\lambda_{\mathrm{BRM}}$. The conceptual distinction from triplet loss is directional: triplet loss pulls features toward the correct prototype, whereas BRM additionally pushes them away from the interpolated boundary proxy where they must not lie.

\input{tables/person-reid}

\subsubsection{Training Strategy \& Inference}
\label{subsubsec:training}

We adopt a two-stage optimization to address the initialization discrepancy between the pre-trained backbone and the randomly initialized prototypes. 

\noindent \textbf{Prototype Warmup.} With the backbone frozen, we optimize only the prototypes via $\mathcal{L}_{\text{w}} = \lambda_{\text{fpt}}\mathcal{L}_{\text{fpt}} + \lambda_{\text{cen}}\mathcal{L}_{\text{cen}} + \lambda_{\text{norm}}\mathcal{L}_{\text{norm}}$, using high learning rates and direct (non-EMA) updates to drive them to stable identity-class anchors within the fixed feature space. We disable $\mathcal{L}_{\text{BRM}}$, whose boundary-proxy construction requires a settled prototype geometry, and $\mathcal{L}_{\text{bht}}$, whose feature-space mining cannot reshape a frozen backbone.

\noindent \textbf{Joint Fine-tuning.} We unfreeze the backbone and train end-to-end with separate learning rates, conservative for the backbone to preserve pre-trained knowledge and higher for the prototypes to track the evolving feature distribution. To avoid abrupt shifts in feature and prototype geometry, the prototypes follow an Exponential Moving Average (EMA) update:
\begin{equation}
   \mathbf{p}_k^{(t+1)} \leftarrow \gamma_{\text{ema}} \,\hat{\mathbf{p}}_k^{(t)}
+ (1 - \gamma_{\text{ema}}) \,\hat{\mathbf{p}}_k^{(t+1)} ,
\end{equation}
where $\gamma_{\text{ema}}$ is the decay rate, $\hat{\mathbf{p}}_k^{(t)}$ and $\hat{\mathbf{p}}_k^{(t+1)}$ are the current and gradient-updated normalized prototypes, and the resulting prototype $\mathbf{p}_k^{(t+1)}$ is re-normalized to $\mathbb{S}^{d-1}$. We activate $\mathcal{L}_{\text{BRM}}$ and $\mathcal{L}_{\text{bht}}$ at this stage. The prototype geometry settled during warmup ensures boundary repulsion targets genuinely ambiguous regions, while the now-trainable backbone lets the instance triplet compress the residual cross-camera spread.

\noindent \textbf{Inference.} Prototypes are discarded at inference and re-identification reduces to nearest-neighbour retrieval in the normalized embedding space:
\begin{equation}
    \hat{y} = \arg\max_{i} \; \hat{\mathbf{f}}_q^\top \hat{\mathbf{f}}_{g,i},
\end{equation}
which requires no gallery prototypes and adds no overhead over a standard ReID pipeline.

%% file: tables/person-reid.tex
\begin{table*}[t]
\centering
\caption{\textbf{Person ReID Results.} Simple \texttt{ViT-B/16} denotes an ImageNet pre-training; $^\ast$pre-trained on LUPerson \cite{fu2020unsupervised}; $^\dagger$CLIP ViT-L/14 used as a distillation teacher with a ResNet-50 student; $^\ddagger$combines DINOv2 ViT-B/14 and CLIP ViT-B/16.}
\label{tab:person-reid_results}
\setlength{\tabcolsep}{4pt}
\resizebox{\textwidth}{!}{
\begin{tabular}{c|l|c|cc|cc|cc|cc|cc}
\hline
\multirow{2}{*}{\textbf{Model}} &
\multirow{2}{*}{\textbf{Backbone}} &
\multirow{2}{*}{\textbf{\#Params}} &
\multicolumn{2}{c|}{\textbf{Market-1501}} &
\multicolumn{2}{c|}{\textbf{DukeMTMC}} &
\multicolumn{2}{c|}{\textbf{MSMT17}} &
\multicolumn{2}{c|}{\textbf{CUHK03-NP}} &
\multicolumn{2}{c}{\textbf{IUST}} \\
& & & mAP & Rank-1 & mAP & Rank-1 & mAP & Rank-1 & mAP & Rank-1 & mAP & Rank-1 \\
\hline
AdaSP \cite{zhou2023adaptive} & ResNet-50 & 25M & 89.8 & 95.5 & 83.0 & 91.7 & 67.1 & 85.5 & 80.1 & 82.0 & - & - \\
ProNet++ \cite{wang2023rethinking} & ResNet-50 & 25M & 90.2 & 96.0 & - & - & 65.5 & 85.4 & 82.7 & 85.2 & - & - \\
CA-Jaccard \cite{Chen_2024_CVPR} & ResNet-50 & 25M & 86.1 & 94.4 & - & - & 44.3 & 75.1 & - & - & - & - \\
TMGF \cite{li2023transformer-wacv} & ViT-S/16 & 21M & 91.9 & 96.3 & 83.1 & 92.3 & 70.3 & 88.2 & - & - & - & - \\
TransReID \cite{He_2021_ICCV} & ViT-B/16 & 86M & 89.5 & 95.2 & 82.6 & 90.7 & 69.4 & 86.2 & 72.9 & 75.1 & - & - \\
DCAL \cite{zhu2022dual} & ViT-B/16 & 86M & 87.5 & 94.7 & 80.1 & 89.0 & 64.0 & 83.1 & - & - & - & - \\
AAFormer \cite{AAformer} & ViT-B/16 & 86M & 88.0 & 95.4 & 80.9 & 90.1 & 65.6 & 84.4 & 80.3 & 79.0 & - & - \\
PHA \cite{zhang2023pha} & ViT-B/16 & 86M & 90.2 & 96.1 & - & - & 68.9 & 86.1 & 80.3 & 83.2 & - & - \\
PASS \cite{zhu2022pass} & ViT-B/16$^\ast$ & 86M & 93.3 & 96.9 & - & - & 74.3 & 89.7 & - & - & - & - \\
SOLIDER \cite{chen2023beyond} & Swin-B & 88M & \underline{93.9} & 96.9 & - & - & 77.1 & 90.7 & - & - & 42.4 & 51.8 \\
CLIP-ReID \cite{li2022clip} & ViT-B/16 (CLIP) & 86M & 90.5 & 95.4 & 83.1 & 90.8 & 75.8 & 89.7 & 81.6 & 80.9 & \underline{51.6} & \underline{61.4} \\
CLIMB-ReID \cite{climb} & ViT-B/16 (CLIP) & 86M & 92.6 & 96.8 & - & - & 77.8 & 90.5 & - & - & - & - \\
PCL-CLIP \cite{li2023prototypical} & ViT-B/16 (CLIP) & 86M & 91.4 & 95.9 & - & - & 76.1 & 89.8 & - & - & - & - \\
HPL \cite{zhou2025hierarchical} & ViT-B/16 (CLIP) & 86M & 89.8 & 95.9 & 82.9 & 90.3 & \underline{79.0} & 91.0 & - & - & - & - \\
FlexiReID \cite{sun2025flexireid} & ViT-B/16 (CLIP) & 86M & 92.1 & 96.0 & - & - & 67.5 & 83.7 & - & - & - & - \\
CLIP3DReID \cite{liu2024distilling} & ViT-L/14 (CLIP)$^\dagger$ & 304M & 88.4 & 95.6 & - & - & 61.2 & 81.5 & - & - & - & - \\
VICP \cite{zhizhong2025vicp} & ViT-S/14 (DiNOv2) & 21M & 90.3 & 95.5 & - & - & 75.3 & 89.2 & - & - & - & - \\
DRFormer \cite{shu2026drformer} & ViT-B/14 (DiNOv2)$^\ddagger$ & 86M & 92.9 & \underline{97.0} & \underline{85.5} & \underline{92.5} & 78.7 & \textbf{91.3} & \underline{88.1} & \underline{89.6} & - & - \\
\textbf{PBR-ReID (Ours)} & \textbf{ViT-B/16 (DiNOv3)} & 86M & \textbf{94.0} & \textbf{97.1} & \textbf{86.1} & \textbf{93.1} & \textbf{79.7} & \underline{91.2} & \textbf{88.4} & \textbf{89.8} & \textbf{55.2} & \textbf{63.8} \\
\hline
\end{tabular}}
\end{table*}

%% file: sec/4_experiments.tex
\section{Experiments}
\label{sec:exp}

\subsection{Experimental settings}

\noindent \textbf{Datasets and evaluation protocol.} We evaluate on five person ReID benchmarks. Market-1501 \cite{market1501} provides 1,501 identities across 6 cameras (12,936 train /19,732 test images). DukeMTMC \cite{dukemtmc} covers 1,812 identities from 8 campus cameras (16,522 train /17,661 test images). MSMT17 \cite{msmt17} is the most challenging one, with 4,101 identities and 2,621 train / 93,820 test images from 15 cameras under diverse conditions. CUHK03 \cite{cuhk03} provides 1,467 identities (7,365/5,328 images) and is evaluated under the new protocol and IUST \cite{iust} is a recent culturally-aware benchmark with 1,847 identities and 117,455 images. For vehicle ReID, we evaluate our method on VeRi-776 \cite{veri776} which contains 776 identities from 20 urban cameras and VehicleID \cite{vehicleid} that is a large-scale benchmark with 26,267 identities and 221,763 images with multiple test splits of increasing difficulty. Following standard conventions, evaluation is based on Cumulative Matching Characteristic (CMC) curves and mean Average Precision (mAP).

\noindent \textbf{Implementation details.} We use DiNOv3 ViT-B/16 as backbone encoder. Training images are augmented with random horizontal flipping, zero-padding followed by random cropping and random erasing. Each mini-batch is formed by class-balanced sampling of $K' = 8$ identities with $N_k = 4$ instances each, giving a batch size $B = 32$. We adopt a dual-optimizer strategy with AdamW~\cite{loshchilov2017decoupled} (weight decay $0.01$). The backbone uses a cosine-decayed learning rate ($1\mathrm{e}{-5}\rightarrow 1\mathrm{e}{-7}$) and the prototypes a step-decayed schedule (initial $1\mathrm{e}{-1}$, step size $15$). Prototypes are initialized with Xavier initialization (gain $0.5$). Training runs for $200$ epochs on a single RTX4090 GPU, with the first $10$ epochs as prototype warmup; during joint fine-tuning the prototypes are updated by EMA with decay $\gamma_{\text{ema}} = 0.999$. Loss weights are $\lambda_{\mathrm{fpt}} = 1.0$, $\lambda_{\mathrm{bht}} = 1.0$, $\lambda_{\mathrm{cen}} = 0.5$, $\lambda_{\mathrm{norm}} = 0.005$, and $\lambda_{\mathrm{BRM}} = 0.4$. Feature-prototype triplet uses margin $m_{\mathrm{fpt}} = 0.3$ and adaptive-weight strength $\gamma_{\text{fpt}} = 2.0$, batch-hard feature triplet uses margin $m_{\mathrm{bht}} = 0.3$, center loss uses similarity-weight strength $\gamma_{\text{center}}^{\text{sim}} = 1.0$ and variance-weight strength $\gamma_{\text{center}}^{\text{var}} = 2.5$. For BRM, we set $\tau = 0.7$, $\alpha = 0.4$, $m_{\mathrm{BRM}} = 0.5$ with repulsion strength $\beta = 0.7$, and $\gamma_{\mathrm{sep}} = 1.5$; the intra-class spread $\hat{\sigma}_k$ is tracked by an EMA with decay $0.9$. At inference we use single-query cosine nearest-neighbor retrieval.

\subsection{State-of-the-art comparison}

\input{tables/vehicle-reid}

All results are reported without re-ranking and camera-aware SIE embeddings.

\noindent\textbf{Person ReID.} Table~\ref{tab:person-reid_results} reports PBR-ReID results on five person ReID benchmarks. To keep the comparison meaningful, we evaluate against methods operating at different design approaches, backbones and pre-training strategies, while comparable at parameter count. Overall, our method achieves strong performance across all benchmarks, especially regarding mAP where it consistently outperforms other approaches. In detail, PBR-ReID yields higher mAP on MSMT17, which is the most challenging benchmark, by +0.7\% compared to HPL \cite{zhou2025hierarchical} and competitive rank-1 to DRFormer \cite{shu2026drformer} trailing by only 0.1\%. PBR-ReID outperforms all others compared for both mAP and rank-1 metrics on Market-1501, DukeMTMC, CUHK03-NP (detected, new protocol) and IUST, even against ones that rely on pre-training over large person datasets like LUPerson \cite{fu2020unsupervised}. 

\noindent\textbf{Vehicle ReID.} To demonstrate effective generalization beyond a single domain for re-identification, Table~\ref{tab:vehicle-reid_results} reports PBR-ReID results on two vehicle ReID benchmarks. Our method transfers without modifications and again ranks first, reaching $88.0$ mAP and $98.1$ rank-1 on VeRi-776 and $90.2$ rank-1 on VehicleID (Large split), outperforming both specialized vehicle ReID methods like VehicleNet \cite{vehicle-net} and VehicleMAE \cite{Wang_2025_ICCV} or general ones like CLIP-ReID \cite{li2022clip}. 

\subsection{Ablation studies}

\noindent \textbf{Different backbones.} Table~\ref{tab:ablation_backbones} reports performance replacing the backbone while all training and inference settings are fixed. Our method remains strong across all encoders, confirming that the gains stem from our formulation rather than from a single backbone. Performance tracks the quality of the pre-trained representation with self-supervised backbones outperforming the CLIP and supervised variants. We attribute this to their semantically richer embedding space since prototypes anchored where identities are already well separated converge to more discriminative positions. Each variant surpasses the best competitor in Table~\ref{tab:person-reid_results} built on the same backbone family. On Market-1501, our CLIP ViT-B/16 variant exceeds the strongest CLIP-based method (CLIMB-ReID \cite{climb}) by $0.8\%$ mAP, our supervised ViT-B/16 variant exceeds the strongest ImageNet-pretrained ViT-B competitor (PHA \cite{zhang2023pha}) by $2.0\%$ mAP, and our DINOv2 ViT-B/14 variant exceeds the strongest DINOv2-based competitor (DRFormer \cite{shu2026drformer}) by $0.8\%$ mAP. The same ordering holds on DukeMTMC. 

\input{tables/ablations_backbones}

\noindent\textbf{Comparison with prototype-based objectives.} To establish a set of baselines, we compare against widely used prototype-based objectives, namely ArcFace \cite{deng2019arcface}, CosFace \cite{wang2018cosface}, CircleLoss \cite{sun2020circle} and VPL \cite{deng2021variational}. We train each head on the same DINOv3 ViT-B/16 backbone under an identical protocol. PBR-ReID improves over the strongest baseline by $3.0\%$ mAP on Market-1501 and $3.5\%$ on DukeMTMC (Table ~\ref{tab:baselines_comparison}).

\input{tables/baselines_comparison}

\noindent \textbf{Loss terms.} Table~\ref{tab:ablation_losses} isolates the contribution of each loss term with experiments on Market-1501 and DukeMTMC. The instance-level batch-hard triplet alone is already a strong baseline, but introducing the prototypes does not immediately show performance gains since the feature-prototype triplet term lowers mAP and adding the center loss recovers only part of the gap on both datasets. This is expected, as the prototypes are an additional set of parameters that must be co-optimized with the backbone and remain unstable on the hypersphere before their geometry settles. However, introducing normalization resolves this and lifts mAP performance by $+4.9\%$ on Market-1501 and by $+3.9\%$ on DukeMTMC, thus establishing that constraining prototypes and embeddings to the unit hypersphere is a prerequisite for stable joint optimization. Once this constraint is in place, the prototype-based configuration surpasses the instance-level baseline, reaching $92.4\%$ mAP on Market-1501 and $84.6\%$ on DukeMTMC. 
Also, detaching the prototype from $\mathcal{L}_{\mathrm{cen}}$ could naturally move only the features toward it, but on Market-1501 and DukeMTMC this lowered mAP by 1-1.5\%, so we keep gradients on both. The coupling matters, since co-adaptation keeps each prototype a reachable anchor, whereas a frozen one becomes a fixed target lying in low-density regions of multi-modal classes.

\input{tables/ablations_losses}

\input{tables/ablations-joint}

\noindent \textbf{BRM effectiveness.} Table~\ref{tab:ablation_losses} also isolates the role of BRM. Standard MixUp brings a marginal improvement over the prototype baseline ($+0.5\%$ mAP on Market-1501, $+0.3\%$ on DukeMTMC), but using neighbor-guided MixUp which interpolates the most similar identity pairs is actively harmful and drops performance even below the no-MixUp baseline. This confirms our premise that synthesizing samples inside the most confusable boundary regions blurs the decision boundary rather than sharpening it. BRM operates on exactly the same confusable pairs as neighbor-guided MixUp but inverts the operation, repelling features away from the interpolated boundary-proxy instead of attracting them toward it. This single change reverses the effect and yields the largest gain of the whole framework, $+1.6\%$ mAP on Market-1501 and $+1.5\%$ on DukeMTMC over the baseline, and $+2.3\%$ and $+1.9\%$ over neighbor-guided MixUp. Since pair selection is held fixed across the two variants, the improvement is attributable to the repulsion mechanism itself rather than to which identities are mixed. 

To further substantiate the contribution of BRM, we evaluate its impact on disambiguating visually similar identities on the Market-1501 test set. We define ``confusable pairs'' as image pairs with an initial cosine similarity $\ge 0.7$. By tracking the prototype distance ($d = 1 - S$) between these specific pairs from an early pre-BRM checkpoint (Epoch 20) to the final model (Epoch 190), we quantify the evolution of the embedding space. As illustrated in Figure~\ref{fig:ablations}, the Kernel Density Estimate (KDE) of the inter-class distances shows a clear rightward shift over the course of training. The mean distance between confusable identities increases from $0.2393$ to $0.2514$, a $5.1\%$ improvement in feature separation. Our training process consistently pushes hard-to-distinguish samples apart, even when they occupy high-density regions of the feature manifold.

\noindent\textbf{Impact of training framework.} Without warmup, mAP collapses to $51.4\%$ on Market-1501 and $50.9\%$ on DukeMTMC, confirming that the randomly initialized prototypes must be decoupled from the backbone before joint training (Figure ~\ref{fig:ablations}). Performance peaks at $10$ epochs and stays stable for longer schedules, indicating low sensitivity to this setting. Also, our experiments have shown that removing the EMA prototype update lowers mAP by 12.3\% on Market-1501 and 11.1\% on DukeMTMC, while replacing the dual learning-rate schedule with a single rate costs a further 2.6\% and 2.2\%. Further training configurations are reported in the supplementary material.

\noindent \textbf{Hyperparameter sensitivity ablation.} Given the
multi-component nature of our training objective, we assess whether the reported gains are contingent on precise hyperparameter tuning. Across the four hyperparameters that govern BRM and the adaptive prototype separation, mAP stays within $1.1$ points of the $94.0\%$ peak (Fig.~\ref{fig:ablations}). The bounded BRM parameters $\lambda_{\mathrm{BRM}}$, $\tau$ and $\beta$, swept over their entire $[0,1]$ domain, vary by at most $1.1$, $0.9$ and $1.1$ points respectively, while $\gamma_{\mathrm{sep}}$ is the most stable at $0.7$. The lower end of $\lambda_{\mathrm{BRM}}$ corresponds to disabling BRM, so even the largest perturbation degrades gracefully rather than collapsing the framework, confirming that PBR-ReID does not rely on delicate tuning to reach its reported accuracy.

%% file: tables/vehicle-reid.tex
\begin{table}[t]
\centering
\caption{\textbf{Vehicle ReID Results.}}
\label{tab:vehicle-reid_results}
\setlength{\tabcolsep}{4pt}
\begin{tabular}{c|cc|cc}
\hline
\multirow{2}{*}{\textbf{Model}} &
\multicolumn{2}{c|}{\textbf{VeRi-776}} &
\multicolumn{2}{c|}{\textbf{VehicleID}}
 \\
& mAP & Rank-1 & Rank-1 & Rank-5  \\
\hline
CFA-Net \cite{cfa-net} & 80.7 & 96.9 & - & - \\
APC \cite{wang2025makes} & 85.1 & \underline{97.9} & - & - \\
VehicleNet \cite{vehicle-net} & 83.4 & 96.8 & 79.5 & 92.0 \\ 
TANet \cite{hu2024tanet} & 83.6 & 96.8 & 78.2 & 92.6 \\
HCI-Net \cite{sun2024heterogeneous} & 83.8 & 96.6 & 76.4 & 91.2 \\
CAN \cite{can} & 83.4 & 97.6 & 80.5 & 94.5 \\
CLIP-ReID \cite{li2022clip} & 84.5 & 97.3 & 85.5 & \underline{97.2} \\
ADPRP-Net \cite{zhou2025adaptive} & 82.8 & 95.6 & 76.7 & 91.9 \\
VehicleMAE \cite{Wang_2025_ICCV} & \underline{87.6} & 97.4 & \underline{85.9} & 94.9 \\
TransReID \cite{He_2021_ICCV} & 82.0 & 97.1 & 85.2 & 97.1 \\ 
AdaSP \cite{zhou2023adaptive} & 82.7 & 96.7 & 83.0 & 94.9 \\
PCL-CLIP \cite{li2023prototypical} & 82.5 & 97.1 & - & - \\
\textbf{PBR-ReID (Ours)} & \textbf{88.0} & \textbf{98.1} & \textbf{90.2} & \textbf{96.0}\\
\hline
\end{tabular}
\end{table}

%% file: tables/ablations_backbones.tex
\begin{table}[t]
  \centering
  \caption{\textbf{Effect of the backbone encoder.} We evaluate PBR-ReID with different pre-trained ViT-B/16 backbones on Market1501~\cite{market1501} and DukeMTMC~\cite{dukemtmc}. All variants use identical training and inference settings.}
  \label{tab:ablation_backbones}
  \setlength{\tabcolsep}{4pt}
  \footnotesize
  \begin{tabular}{l cc cc}
    \toprule
    & \multicolumn{2}{c}{Market-1501} & \multicolumn{2}{c}{DukeMTMC} \\
    \cmidrule(lr){2-3} \cmidrule(lr){4-5}
    Backbone & mAP & Rank-1 & mAP & Rank-1 \\
    \midrule
    ViT-B/16 (ImageNet)        & 92.2 & 96.3 & 84.1 & 91.8 \\
    CLIP ViT-B/16           & 93.4 & 96.6 & 85.4 & 92.7 \\
    DINOv2 ViT-B/14                 & 93.7 & 96.7 & 85.9 & 92.9 \\
    DINOv3 ViT-B/16           & \textbf{94.0} & \textbf{97.1} & \textbf{86.1} & \textbf{93.1} \\
    \bottomrule
  \end{tabular}
\end{table}

%% file: tables/baselines_comparison.tex
\begin{table}[t]
  \centering
  \caption{\textbf{Comparison with prototype-based objectives.} All variants
  share the same DINOv3 ViT-B/16 backbone and training protocol, only the
  prototype-based objective differs.}
  \label{tab:baselines_comparison}
  \setlength{\tabcolsep}{4pt}
  \footnotesize
  \begin{tabular}{l cc cc}
    \toprule
    & \multicolumn{2}{c}{Market-1501} & \multicolumn{2}{c}{DukeMTMC} \\
    \cmidrule(lr){2-3} \cmidrule(lr){4-5}
    Training Objective & mAP & Rank-1 & mAP & Rank-1 \\
    \midrule
    ArcFace \cite{deng2019arcface}    & 90.2 & 94.9 & 82.0 & 90.7 \\
    CosFace \cite{wang2018cosface}    & 89.6 & 95.8 & 82.6 & \underline{92.4} \\
    CircleLoss \cite{sun2020circle}   & \underline{91.0} & \underline{96.0} & \underline{83.9} & \underline{92.4} \\ 
    VPL \cite{deng2021variational}    & 90.0 & 95.0 & 81.9 & 90.6 \\
    \midrule
    \textbf{PBR-ReID (Ours)}          & \textbf{94.0} & \textbf{97.1} & \textbf{86.1} & \textbf{93.1} \\
    \bottomrule
  \end{tabular}
\end{table}

%% file: tables/ablations_losses.tex
\begin{table}[t]
  \centering
  \caption{\textbf{Effect of loss terms \& BRM.} The top block uses instance-level supervision only (no prototypes); the bottom block introduces them. NG: neighbor-guided MixUp.}
  \label{tab:ablation_losses}
  \setlength{\tabcolsep}{3.5pt}
  \resizebox{\columnwidth}{!}{%
  \begin{tabular}{cccccc cc cc}
    \toprule
    \multirow{2}{*}{$\mathcal{L}_{\mathrm{bht}}$} & \multirow{2}{*}{$\mathcal{L}_{\mathrm{fpt}}$}
    & \multirow{2}{*}{$\mathcal{L}_{\mathrm{cen}}$} & \multirow{2}{*}{$\mathcal{L}_{\mathrm{norm}}$}
    & \multirow{2}{*}{MixUp} & \multirow{2}{*}{BRM}
    & \multicolumn{2}{c}{Market-1501} & \multicolumn{2}{c}{DukeMTMC} \\
    \cmidrule(lr){7-8} \cmidrule(lr){9-10}
    & & & & & & mAP & Rank-1 & mAP & Rank-1 \\
    \midrule
    \checkmark &            &            &            &            &            & 88.9 & 94.5 &   82.1   & 90.8     \\
    \checkmark &            &            & \checkmark &            &            & 90.4 & 95.2 &   83.2   & 91.3     \\
    \midrule
    \checkmark & \checkmark &            &            &            &            & 86.3 & 93.0 & 79.8 & 88.7 \\
    \checkmark & \checkmark & \checkmark &            &            &            & 87.5 & 93.7 & 80.7 & 89.9 \\
    \checkmark & \checkmark & \checkmark & \checkmark &            &            & 92.4 & 96.3 & 84.6 & 92.4 \\
    \checkmark & \checkmark & \checkmark & \checkmark & \checkmark &            & 92.9 & 96.6 & 84.9 & 92.6 \\
    \checkmark & \checkmark & \checkmark & \checkmark & \checkmark (NG) &       & 91.7 & 95.9 & 84.2 & 92.1 \\
    \checkmark & \checkmark & \checkmark & \checkmark &  \checkmark          & \checkmark & \textbf{94.0} & \textbf{97.1} & \textbf{86.1} & \textbf{93.1} \\
    \bottomrule
  \end{tabular}%
  }
\end{table}

%% file: tables/ablations-joint.tex
\begin{figure*}[t]
  \centering
  \resizebox{\textwidth}{!}{%
    \begin{subfigure}[b]{0.50\textwidth}
      \centering
      \input{tables/ablations_sensitivity}
      \label{fig:combined_sensitivity}
    \end{subfigure}
    
    \hfill %
    
    \begin{subfigure}[b]{0.42\textwidth}
      \centering
      \input{tables/warmup}
      \label{fig:warmup_ablation}
    \end{subfigure}

  \hfill

    \begin{subfigure}[b]{0.70\textwidth}
      \centering
      \input{tables/kde_distribution_market1501}
    \label{fig:kde}
    \end{subfigure}
  }
  \caption{\textbf{Ablation studies.} Hyperparameter ablations (left) and KDEs of inter-class feature distances on Market-1501 (right), warmup epochs experiments (center)}
  \label{fig:ablations}
\end{figure*}
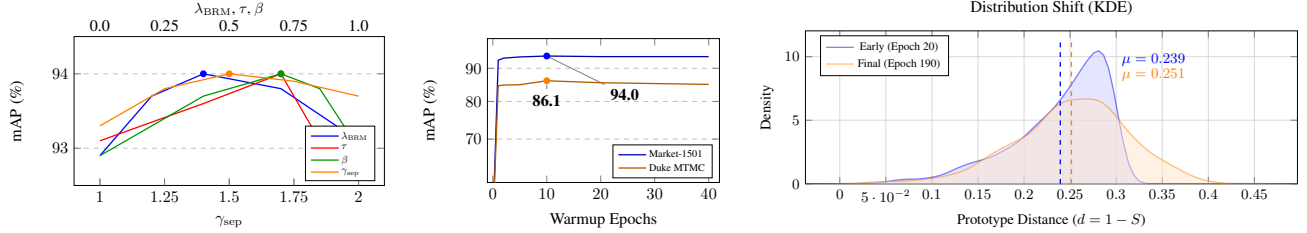

%% file: tables/ablations_sensitivity.tex
\label{fig:ablation-sens}
\centering
\begin{tikzpicture}
\begin{axis}[
    width=0.95\linewidth,
    height=0.55\linewidth,
    xlabel={$\gamma_{\mathrm{sep}}$},
    ylabel={mAP (\%)},
    xtick pos=left,
    xtick={1.0, 1.25, 1.5, 1.75, 2.0},
    extra x ticks={1.0, 1.25, 1.5, 1.75, 2.0},
    extra x tick labels={0.0, 0.25, 0.5, 0.75, 1.0},
    extra x tick style={
        xticklabel pos=upper,
        grid=none
    },
    ymin=92.5, ymax=94.5,
    ytick={93.0, 94.0},
    ymajorgrids=true,
    grid style=dashed,
    legend pos=south east,
    legend style={
        font=\small,
        cells={anchor=west},
        nodes={scale=0.7, transform shape}
    },
    mark size=2pt,
    clip=false
]
\node[anchor=south, font=\small] at (axis description cs:0.5, 1.1)
    {$\lambda_{\mathrm{BRM}}$, $\tau$, $\beta$};

\addplot[color=blue, thick] coordinates {
    (1.0, 92.9) (1.2, 93.7) (1.4, 94.0) (1.7, 93.8) (2.0, 93.1)
};
\node[fill=blue, circle, inner sep=1.5pt] at (axis cs:1.4, 94.0) {};
\addlegendentry{$\lambda_{\mathrm{BRM}}$}

\addplot[color=red, thick] coordinates {
    (1.0, 93.1) (1.4, 93.6) (1.7, 94.0) (1.85, 93.1) (2.0, 93.2)
};
\node[fill=red, circle, inner sep=1.5pt] at (axis cs:1.7, 94.0) {};
\addlegendentry{$\tau$}

\addplot[color=green!60!black, thick] coordinates {
    (1.0, 92.9) (1.4, 93.7) (1.7, 94.0) (1.85, 93.8) (2.0, 93.0)
};
\node[fill=green!60!black, circle, inner sep=1.5pt] at (axis cs:1.7, 94.0) {};
\addlegendentry{$\beta$}

\addplot[color=orange, thick] coordinates {
    (1.0, 93.3) (1.25, 93.8) (1.5, 94.0) (1.75, 93.9) (2.0, 93.7)
};
\node[fill=orange, circle, inner sep=1.5pt] at (axis cs:1.5, 94.0) {};
\addlegendentry{$\gamma_{\mathrm{sep}}$}

\end{axis}
\end{tikzpicture}

%% file: tables/warmup.tex
\centering
\begin{tikzpicture}
\begin{axis}[
    width=0.9\linewidth,
    height=0.64\linewidth,
    xlabel={Warmup Epochs},
    ylabel={mAP (\%)},
    xtick={0,10, 20,30,40},
    xmin=-1, xmax=42,
    ymode=log,
    log ticks with fixed point,
    ymin=60, ymax=100,
    ytick={70, 80, 90},
    ymajorgrids=true,
    grid style=dashed,
    mark size=2.5pt,
    legend pos=south east,
    legend style={
        font=\small,           %
        cells={anchor=west},   %
        nodes={scale=0.7, transform shape} %
    },
]

\addplot[
    color=blue!70!black,
    mark=circle*,
    thick,
    ]
    coordinates {
    (0, 51.4) (1, 92.6) (2, 93.3) (5, 93.7) (10, 94.0) (20, 93.8) (40, 93.8)
    };

\node[pin={[pin distance=1.3cm]335:{\textbf{94.0}}}, fill=blue, circle, inner sep=1.5pt]
    at (axis cs:10, 94.0) {};

\addlegendentry{Market-1501}

\addplot[
    color=orange!70!black,
    mark=circle*,
    thick,
    ]
    coordinates {
    (0, 50.9) (1, 84.6) (2, 84.8) (5, 84.9) (10, 86.1) (20, 85.5) (40, 85.0)
    };

\node[pin={[pin distance=0.1cm]270:{\textbf{86.1}}}, fill=orange, circle, inner sep=1.5pt]
    at (axis cs:10, 86.1) {};

\addlegendentry{Duke MTMC}
    
\end{axis}
\end{tikzpicture}

%% file: tables/kde_distribution_market1501.tex
\begin{tikzpicture}
\begin{axis}[
  width=\linewidth,
  height=0.4\linewidth,
  xmin=-0.0373, xmax=0.4967,
  ymin=0, ymax=12.0296,
  xlabel={Prototype Distance ($d = 1 - S$)},
  ylabel={Density},
  title={Distribution Shift (KDE)},
  grid=both,
  grid style={line width=0.2pt, draw=gray!30},
  legend pos=north west,
  legend style={font=\scriptsize},
  ylabel near ticks,
  tick label style={font=\small},
  label style={font=\small},
]

\addplot[fill=blue!20, fill opacity=0.5, draw=none, forget plot]
  coordinates {(-0.037321,0.000000) (-0.031928,0.000000) (-0.026534,0.000002) (-0.021140,0.000012) (-0.015746,0.000064) (-0.010352,0.000275) (-0.004959,0.000957) (0.000435,0.002706) (0.005829,0.006239) (0.011223,0.011804) (0.016617,0.018655) (0.022010,0.025632) (0.027404,0.033078) (0.032798,0.044266) (0.038192,0.064387) (0.043586,0.097838) (0.048979,0.146074) (0.054373,0.206158) (0.059767,0.269092) (0.065161,0.320979) (0.070555,0.351224) (0.075948,0.362762) (0.081342,0.371578) (0.086736,0.393242) (0.092130,0.431780) (0.097523,0.484129) (0.102917,0.552931) (0.108311,0.649791) (0.113705,0.783952) (0.119099,0.950058) (0.124492,1.129428) (0.129886,1.302441) (0.135280,1.458046) (0.140674,1.593695) (0.146068,1.711837) (0.151461,1.819569) (0.156855,1.929787) (0.162249,2.059056) (0.167643,2.220769) (0.173037,2.416716) (0.178430,2.634951) (0.183824,2.861595) (0.189218,3.099875) (0.194612,3.373421) (0.200005,3.702296) (0.205399,4.076297) (0.210793,4.460516) (0.216187,4.829346) (0.221581,5.188989) (0.226974,5.566658) (0.232368,5.985762) (0.237762,6.453551) (0.243156,6.965612) (0.248550,7.515169) (0.253943,8.095528) (0.259337,8.694934) (0.264731,9.291819) (0.270125,9.848967) (0.275519,10.290061) (0.280912,10.460529) (0.286306,10.127877) (0.291700,9.088917) (0.297094,7.352023) (0.302488,5.229939) (0.307881,3.207779) (0.313275,1.671191) (0.318669,0.731410) (0.324063,0.266746) (0.329456,0.080588) (0.334850,0.020081) (0.340244,0.004114) (0.345638,0.000691) (0.351032,0.000095) (0.356425,0.000011) (0.361819,0.000001) (0.367213,0.000000) (0.372607,0.000000) (0.378001,0.000000) (0.383394,0.000000) (0.388788,0.000000) (0.394182,0.000000) (0.399576,0.000000) (0.404970,0.000000) (0.410363,0.000000) (0.415757,0.000000) (0.421151,0.000000) (0.426545,0.000000) (0.431939,0.000000) (0.437332,0.000000) (0.442726,0.000000) (0.448120,0.000000) (0.453514,0.000000) (0.458907,0.000000) (0.464301,0.000000) (0.469695,0.000000) (0.475089,0.000000) (0.480483,0.000000) (0.485876,0.000000) (0.491270,0.000000) (0.496664,0.000000)};

\addplot[blue, thick, opacity=0.5]
  coordinates {(-0.037321,0.000000) (-0.031928,0.000000) (-0.026534,0.000002) (-0.021140,0.000012) (-0.015746,0.000064) (-0.010352,0.000275) (-0.004959,0.000957) (0.000435,0.002706) (0.005829,0.006239) (0.011223,0.011804) (0.016617,0.018655) (0.022010,0.025632) (0.027404,0.033078) (0.032798,0.044266) (0.038192,0.064387) (0.043586,0.097838) (0.048979,0.146074) (0.054373,0.206158) (0.059767,0.269092) (0.065161,0.320979) (0.070555,0.351224) (0.075948,0.362762) (0.081342,0.371578) (0.086736,0.393242) (0.092130,0.431780) (0.097523,0.484129) (0.102917,0.552931) (0.108311,0.649791) (0.113705,0.783952) (0.119099,0.950058) (0.124492,1.129428) (0.129886,1.302441) (0.135280,1.458046) (0.140674,1.593695) (0.146068,1.711837) (0.151461,1.819569) (0.156855,1.929787) (0.162249,2.059056) (0.167643,2.220769) (0.173037,2.416716) (0.178430,2.634951) (0.183824,2.861595) (0.189218,3.099875) (0.194612,3.373421) (0.200005,3.702296) (0.205399,4.076297) (0.210793,4.460516) (0.216187,4.829346) (0.221581,5.188989) (0.226974,5.566658) (0.232368,5.985762) (0.237762,6.453551) (0.243156,6.965612) (0.248550,7.515169) (0.253943,8.095528) (0.259337,8.694934) (0.264731,9.291819) (0.270125,9.848967) (0.275519,10.290061) (0.280912,10.460529) (0.286306,10.127877) (0.291700,9.088917) (0.297094,7.352023) (0.302488,5.229939) (0.307881,3.207779) (0.313275,1.671191) (0.318669,0.731410) (0.324063,0.266746) (0.329456,0.080588) (0.334850,0.020081) (0.340244,0.004114) (0.345638,0.000691) (0.351032,0.000095) (0.356425,0.000011) (0.361819,0.000001) (0.367213,0.000000) (0.372607,0.000000) (0.378001,0.000000) (0.383394,0.000000) (0.388788,0.000000) (0.394182,0.000000) (0.399576,0.000000) (0.404970,0.000000) (0.410363,0.000000) (0.415757,0.000000) (0.421151,0.000000) (0.426545,0.000000) (0.431939,0.000000) (0.437332,0.000000) (0.442726,0.000000) (0.448120,0.000000) (0.453514,0.000000) (0.458907,0.000000) (0.464301,0.000000) (0.469695,0.000000) (0.475089,0.000000) (0.480483,0.000000) (0.485876,0.000000) (0.491270,0.000000) (0.496664,0.000000)};
\addlegendentry{Early (Epoch 20)}

\addplot[fill=orange!20, fill opacity=0.5, draw=none, forget plot]
  coordinates {(-0.037321,0.000035) (-0.031928,0.000125) (-0.026534,0.000386) (-0.021140,0.001035) (-0.015746,0.002419) (-0.010352,0.004943) (-0.004959,0.008925) (0.000435,0.014498) (0.005829,0.021811) (0.011223,0.031548) (0.016617,0.045303) (0.022010,0.065029) (0.027404,0.091411) (0.032798,0.122172) (0.038192,0.152016) (0.043586,0.174996) (0.048979,0.188000) (0.054373,0.192800) (0.059767,0.195008) (0.065161,0.200808) (0.070555,0.214084) (0.075948,0.235849) (0.081342,0.265598) (0.086736,0.302849) (0.092130,0.347647) (0.097523,0.400288) (0.102917,0.461024) (0.108311,0.529978) (0.113705,0.607324) (0.119099,0.694159) (0.124492,0.794191) (0.129886,0.914849) (0.135280,1.065641) (0.140674,1.253288) (0.146068,1.476303) (0.151461,1.723273) (0.156855,1.977180) (0.162249,2.223558) (0.167643,2.456544) (0.173037,2.677872) (0.178430,2.890150) (0.183824,3.092202) (0.189218,3.283153) (0.194612,3.472404) (0.200005,3.684218) (0.205399,3.948914) (0.210793,4.285293) (0.216187,4.687544) (0.221581,5.125425) (0.226974,5.555460) (0.232368,5.934959) (0.237762,6.233137) (0.243156,6.437884) (0.248550,6.557644) (0.253943,6.616990) (0.259337,6.645496) (0.264731,6.662959) (0.270125,6.667486) (0.275519,6.633568) (0.280912,6.523175) (0.286306,6.305190) (0.291700,5.972429) (0.297094,5.546438) (0.302488,5.067909) (0.307881,4.579499) (0.313275,4.111675) (0.318669,3.678485) (0.324063,3.282409) (0.329456,2.921996) (0.334850,2.596613) (0.340244,2.306648) (0.345638,2.050597) (0.351032,1.821826) (0.356425,1.608312) (0.361819,1.397332) (0.367213,1.182767) (0.372607,0.969105) (0.378001,0.768206) (0.383394,0.591497) (0.388788,0.444220) (0.394182,0.325422) (0.399576,0.231581) (0.404970,0.159529) (0.410363,0.106769) (0.415757,0.070475) (0.421151,0.047101) (0.426545,0.033021) (0.431939,0.025312) (0.437332,0.021826) (0.442726,0.020707) (0.448120,0.020115) (0.453514,0.018578) (0.458907,0.015543) (0.464301,0.011489) (0.469695,0.007415) (0.475089,0.004157) (0.480483,0.002020) (0.485876,0.000850) (0.491270,0.000309) (0.496664,0.000097)};

\addplot[orange, thick, opacity=0.5]
  coordinates {(-0.037321,0.000035) (-0.031928,0.000125) (-0.026534,0.000386) (-0.021140,0.001035) (-0.015746,0.002419) (-0.010352,0.004943) (-0.004959,0.008925) (0.000435,0.014498) (0.005829,0.021811) (0.011223,0.031548) (0.016617,0.045303) (0.022010,0.065029) (0.027404,0.091411) (0.032798,0.122172) (0.038192,0.152016) (0.043586,0.174996) (0.048979,0.188000) (0.054373,0.192800) (0.059767,0.195008) (0.065161,0.200808) (0.070555,0.214084) (0.075948,0.235849) (0.081342,0.265598) (0.086736,0.302849) (0.092130,0.347647) (0.097523,0.400288) (0.102917,0.461024) (0.108311,0.529978) (0.113705,0.607324) (0.119099,0.694159) (0.124492,0.794191) (0.129886,0.914849) (0.135280,1.065641) (0.140674,1.253288) (0.146068,1.476303) (0.151461,1.723273) (0.156855,1.977180) (0.162249,2.223558) (0.167643,2.456544) (0.173037,2.677872) (0.178430,2.890150) (0.183824,3.092202) (0.189218,3.283153) (0.194612,3.472404) (0.200005,3.684218) (0.205399,3.948914) (0.210793,4.285293) (0.216187,4.687544) (0.221581,5.125425) (0.226974,5.555460) (0.232368,5.934959) (0.237762,6.233137) (0.243156,6.437884) (0.248550,6.557644) (0.253943,6.616990) (0.259337,6.645496) (0.264731,6.662959) (0.270125,6.667486) (0.275519,6.633568) (0.280912,6.523175) (0.286306,6.305190) (0.291700,5.972429) (0.297094,5.546438) (0.302488,5.067909) (0.307881,4.579499) (0.313275,4.111675) (0.318669,3.678485) (0.324063,3.282409) (0.329456,2.921996) (0.334850,2.596613) (0.340244,2.306648) (0.345638,2.050597) (0.351032,1.821826) (0.356425,1.608312) (0.361819,1.397332) (0.367213,1.182767) (0.372607,0.969105) (0.378001,0.768206) (0.383394,0.591497) (0.388788,0.444220) (0.394182,0.325422) (0.399576,0.231581) (0.404970,0.159529) (0.410363,0.106769) (0.415757,0.070475) (0.421151,0.047101) (0.426545,0.033021) (0.431939,0.025312) (0.437332,0.021826) (0.442726,0.020707) (0.448120,0.020115) (0.453514,0.018578) (0.458907,0.015543) (0.464301,0.011489) (0.469695,0.007415) (0.475089,0.004157) (0.480483,0.002020) (0.485876,0.000850) (0.491270,0.000309) (0.496664,0.000097)};
\addlegendentry{Final (Epoch 190)}

\addplot[blue, dashed, thick] coordinates {(0.239331,0) (0.239331,11.5)};
\addplot[orange, dashed, thick] coordinates {(0.251429,0) (0.251429,11.5)};

\node[blue, font=\small\bfseries, anchor=west] at (axis cs:0.30,9.8)
  {$\mu=0.239$};
\node[orange, font=\small\bfseries, anchor=west] at (axis cs:0.30,8.6)
  {$\mu=0.251$};

\end{axis}
\end{tikzpicture}

%% file: sec/5_conclusions.tex
\section{Conclusions}
\label{sec:conclusions}

We presented PBR-ReID, a prototype-based framework that makes inter-class
geometry an explicit learning target for object re-identification. Building on a learnable prototype per identity, we introduced Boundary Repulsion MixUp (BRM), which reads the most confusable identity pairs from the prototype similarity matrix and repels their features from an interpolated boundary proxy, turning the region standard MixUp blurs into a source of discrimination. With a DiNOv3 ViT-B/16 backbone, PBR-ReID achieved strong results across competitive benchmarks. Our ablations confirmed that the gains originate from the method rather than the backbone and that performance is robust to its hyperparameters. A promising direction is to extend our prototype-based boundary repulsion to unsupervised ReID, where prototypes are derived from cluster pseudo-labels rather than ground-truth identities.

%% file: sec/00_supplementary.tex
\section{Additional Experiments}
\suppressfloats[t]

\noindent \textbf{Performance scaling \& Efficiency.} 
Since the learnable prototypes and BRM act only during training and are dropped before retrieval, inference reduces to a single backbone forward pass followed by nearest-neighbour matching, so the test-time cost is set entirely by the chosen encoder and matches that of a plain ViT of the same size. The prototype matrix, central to structuring the training-time geometry and to BRM, therefore adds nothing at test time. This decoupling lets us trade accuracy for compute purely by swapping the backbone. Table~\ref{tab:scaling} reports PBR-ReID with DINOv3 ViT-B/16, ViT-L/16 and ViT-H+/16, where accuracy improves with capacity while our default ViT-B/16 already reaches competitive accuracy at a fraction of the parameters.

\begin{table}
  \centering
  \caption{\textbf{Performance scaling with backbone capacity.} PBR-ReID with DINOv3 ViT backbones experiments with fixed training and inference settings.}
  \label{tab:scaling}
  \setlength{\tabcolsep}{4pt}
  \footnotesize
  \begin{tabular}{l c cc cc}
    \toprule
    & & \multicolumn{2}{c}{Market-1501} & \multicolumn{2}{c}{DukeMTMC} \\
    \cmidrule(lr){3-4} \cmidrule(lr){5-6}
    Backbone & \#Params & mAP & Rank-1 & mAP & Rank-1 \\
    \midrule
    ViT-B/16  & 86M  & 94.0 & 97.1 & 86.1 & 93.1 \\
    ViT-L/16  & 300M & 94.8 & 97.2 & 89.7 & 94.6 \\
    ViT-H+/16 & 840M & 95.3 & 97.4 & 90.8 & 95.1 \\
    \bottomrule
  \end{tabular}
\end{table}

\begin{table}
  \centering
  \caption{\textbf{Ablation of training stabilization components.} Effect of the dual learning-rate schedule and the EMA prototype update during joint fine-tuning. The top row is the full PBR-ReID configuration.}
  \label{tab:stabilization}
  \setlength{\tabcolsep}{4pt}
  \footnotesize
  \begin{tabular}{cc cc cc}
    \toprule
    \multirow{2}{*}{Dual LR} & \multirow{2}{*}{EMA Decay}
    & \multicolumn{2}{c}{Market-1501} & \multicolumn{2}{c}{DukeMTMC} \\
    \cmidrule(lr){3-4} \cmidrule(lr){5-6}
    & & mAP & Rank-1 & mAP & Rank-1 \\
    \midrule
    \checkmark & 0.999      & \textbf{94.0} & \textbf{97.1} & \textbf{86.1} & \textbf{93.1} \\
    \checkmark & 0.90       &  92.1    &   95.9   &  84.4    &   91.5   \\
    \checkmark & 0 (No EMA) & 81.7 &   90.8   & 75.0 & 85.2 \\
               & 0.999      & 91.4 & 95.8 & 83.9 & 91.9 \\
    \bottomrule
  \end{tabular}
\end{table}

\noindent \textbf{Complementary ReID results.}
We report additional results on the CUHK03-NP benchmark under the labeled (CUHK03-L) setting of the new protocol~\cite{cuhk03}, together with all three test splits of VehicleID~\cite{vehicleid}, in Table~\ref{tab:reid_results_supplementary}. The labeled split provides cleaner bounding-box annotations and is therefore less challenging, reaching 89.7 mAP and 90.6 Rank-1 against 88.4 and 89.8 on the detected split, confirming the expected gap between the two conditions. On VehicleID, PBR-ReID stays strong and consistent across gallery sizes, leading Rank-1 on every split from 94.0 on Small to 90.2 on Large.

\begin{table*}
\centering
\caption{\textbf{Additional ReID results} on CUHK03~\cite{cuhk03} and VehicleID~\cite{vehicleid} benchmarks.}
\label{tab:reid_results_supplementary}
\resizebox{\textwidth}{!}{%
\begin{tabular}{l cc cc cc cc cc}
\toprule
\multirow{3}{*}{Method}
& \multicolumn{2}{c}{CUHK03-D}
& \multicolumn{2}{c}{CUHK03-L}
& \multicolumn{6}{c}{VehicleID} \\
\cmidrule(lr){2-3} \cmidrule(lr){4-5} \cmidrule(lr){6-11}
& & & & & \multicolumn{2}{c}{Small} & \multicolumn{2}{c}{Medium} & \multicolumn{2}{c}{Large} \\
\cmidrule(lr){6-7} \cmidrule(lr){8-9} \cmidrule(lr){10-11}
& mAP & Rank-1 & mAP & Rank-1 & Rank-1 & Rank-5 & Rank-1 & Rank-5 & Rank-1 & Rank-5 \\
\midrule
CLIP-ReID \cite{li2022clip} & - & - & - & - & 85.3 & 97.6 & 81.0 & 95.0 & 78.1 & 92.7 \\
AdaSP \cite{zhou2023adaptive} & 80.1 & 82.0 & 82.4 & 84.6 & 89.3 & \textbf{98.4} & 85.6 & \textbf{96.7} & 83.0 & 94.9 \\
\textbf{PBR-ReID (Ours)} & \textbf{88.4} & \textbf{89.8} & \textbf{89.7} & \textbf{90.6} & \textbf{94.0} & 97.8 & \textbf{92.1} & 96.6 & \textbf{90.2} & \textbf{96.0} \\
\bottomrule
\end{tabular}%
}
\end{table*}

\noindent \textbf{Experiments on training stabilization.}
PBR-ReID's staged protocol relies on two stabilization mechanisms during joint fine-tuning, a dual learning-rate schedule that applies a conservative rate to the backbone and a higher rate to the prototypes, and an EMA update that smooths prototype trajectories across gradient steps. Both address the initialization discrepancy between the pre-trained DINOv3 encoder and the randomly initialized prototype matrix. Table~\ref{tab:stabilization} ablates each component independently on Market-1501 and DukeMTMC. Replacing the dual schedule with a single shared rate forces an unfavorable trade-off between backbone stability and prototype adaptability and lowers mAP to 91.4\% on Market-1501 and 83.9\% on DukeMTMC. Disabling EMA is far more damaging, since the prototypes then absorb high-frequency mini-batch noise and produce unreliable boundary proxies that degrade BRM's repulsion signal, collapsing mAP to 81.7 and 75.0. A lower decay of 0.90 reintroduces part of this noise and trails the 0.999 setting by 1.9 and 1.7 mAP, confirming that strongly smoothed prototype updates are needed for stable convergence.

\section{Visualizations}

We complement the quantitative results with qualitative top-10 retrieval comparisons that isolate the effect of BRM on disambiguating confusable identities across the person and vehicle domains. The queries shown here are deliberately drawn from the hardest, most confusable cases, where competing identities are nearly indistinguishable in appearance, so that the contribution of BRM is visible in isolation. They are not representative of typical retrieval, which is already strong across all benchmarks as reported in the main paper, but are chosen to expose the boundary regions that BRM is designed to address. For each query we show the ranked list of the baseline without BRM above that of the BRM-trained model.

On MSMT17 (Figures~\ref{fig:retrieval-msmt17-01} and~\ref{fig:retrieval-msmt17-02}), the baseline lists frequently rank visually similar but incorrect identities among the top matches, typically pedestrians sharing clothing style and color. BRM promotes the correct matches into the top ranks and suppresses these near-boundary confusions and the gain persists under strong appearance shifts where the baseline fails. Similar behavior is observed also on VeRi-776 (Figures~\ref{fig:retrieval-page-01}), where the dominant source of confusion shifts to vehicles of identical model and color. The baseline top ranks are dominated by same-type vehicles of the wrong identity, whereas the BRM-trained model recovers the correct matches, indicating that the repulsion mechanism transfers to another domain without modification.

%% file: supp/veri_retrieval_comparison.tex
\begin{figure*}[p]
\resizebox{\textwidth}{!}{%
  \centering

  {\setlength{\tabcolsep}{2pt}%
  \begin{tabular}{m{6mm}ccccccccccc}
   & \textbf{Query} & \textcolor{milestone}{\textbf{R-1}} &  &  &  & \textcolor{milestone}{\textbf{R-5}} &  &  &  &  & \textcolor{milestone}{\textbf{R-10}} \\[2pt]
  \rotatebox{90}{\parbox{20mm}{\centering\footnotesize\textbf{No BRM}}} &
    \begin{tabular}{@{}c@{}}
    \adjincludegraphics[width=\imgw,height=\imgh,cfbox=querycol 2pt 0pt]{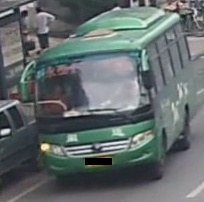}\\[-1pt]
    {\tiny\textbf{\textcolor{querycol}{ID:~66}}}
  \end{tabular} &
    \begin{tabular}{@{}c@{}}
    \adjincludegraphics[width=\imgw,height=\imgh,cfbox=wrong 2pt 0pt]{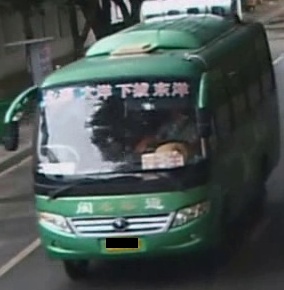}\\[-1pt]
    {\tiny\textbf{\textcolor{wrong}{ID:~65}}}
  \end{tabular} &
    \begin{tabular}{@{}c@{}}
    \adjincludegraphics[width=\imgw,height=\imgh,cfbox=wrong 2pt 0pt]{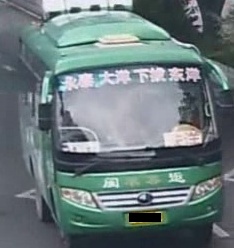}\\[-1pt]
    {\tiny\textbf{\textcolor{wrong}{ID:~65}}}
  \end{tabular} &
    \begin{tabular}{@{}c@{}}
    \adjincludegraphics[width=\imgw,height=\imgh,cfbox=correct 2pt 0pt]{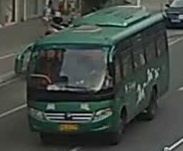}\\[-1pt]
    {\tiny\textbf{\textcolor{correct}{ID:~66}}}
  \end{tabular} &
    \begin{tabular}{@{}c@{}}
    \adjincludegraphics[width=\imgw,height=\imgh,cfbox=wrong 2pt 0pt]{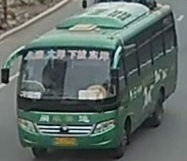}\\[-1pt]
    {\tiny\textbf{\textcolor{wrong}{ID:~65}}}
  \end{tabular} &
    \begin{tabular}{@{}c@{}}
    \adjincludegraphics[width=\imgw,height=\imgh,cfbox=wrong 2pt 0pt]{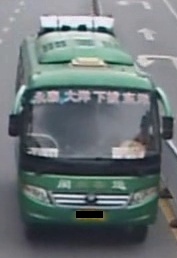}\\[-1pt]
    {\tiny\textbf{\textcolor{wrong}{ID:~65}}}
  \end{tabular} &
    \begin{tabular}{@{}c@{}}
    \adjincludegraphics[width=\imgw,height=\imgh,cfbox=wrong 2pt 0pt]{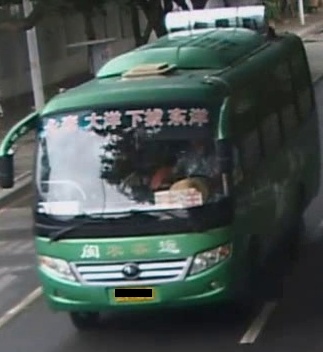}\\[-1pt]
    {\tiny\textbf{\textcolor{wrong}{ID:~65}}}
  \end{tabular} &
    \begin{tabular}{@{}c@{}}
    \adjincludegraphics[width=\imgw,height=\imgh,cfbox=wrong 2pt 0pt]{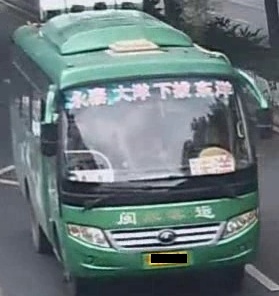}\\[-1pt]
    {\tiny\textbf{\textcolor{wrong}{ID:~65}}}
  \end{tabular} &
    \begin{tabular}{@{}c@{}}
    \adjincludegraphics[width=\imgw,height=\imgh,cfbox=wrong 2pt 0pt]{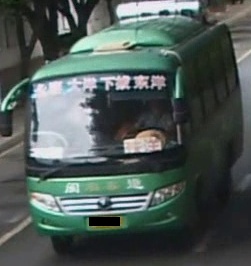}\\[-1pt]
    {\tiny\textbf{\textcolor{wrong}{ID:~65}}}
  \end{tabular} &
    \begin{tabular}{@{}c@{}}
    \adjincludegraphics[width=\imgw,height=\imgh,cfbox=wrong 2pt 0pt]{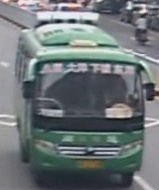}\\[-1pt]
    {\tiny\textbf{\textcolor{wrong}{ID:~65}}}
  \end{tabular} &
    \begin{tabular}{@{}c@{}}
    \adjincludegraphics[width=\imgw,height=\imgh,cfbox=wrong 2pt 0pt]{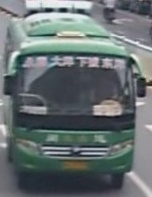}\\[-1pt]
    {\tiny\textbf{\textcolor{wrong}{ID:~65}}}
  \end{tabular} \\[2pt]
  \rotatebox{90}{\parbox{20mm}{\centering\footnotesize\textbf{BRM}}} &
    \begin{tabular}{@{}c@{}}
    \adjincludegraphics[width=\imgw,height=\imgh,cfbox=querycol 2pt 0pt]{supp/images/part3/q01_pid0066.jpg}\\[-1pt]
    {\tiny\textbf{\textcolor{querycol}{ID:~66}}}
  \end{tabular} &
    \begin{tabular}{@{}c@{}}
    \adjincludegraphics[width=\imgw,height=\imgh,cfbox=correct 2pt 0pt]{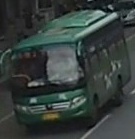}\\[-1pt]
    {\tiny\textbf{\textcolor{correct}{ID:~66}}}
  \end{tabular} &
    \begin{tabular}{@{}c@{}}
    \adjincludegraphics[width=\imgw,height=\imgh,cfbox=correct 2pt 0pt]{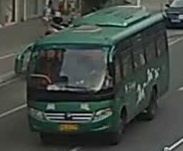}\\[-1pt]
    {\tiny\textbf{\textcolor{correct}{ID:~66}}}
  \end{tabular} &
    \begin{tabular}{@{}c@{}}
    \adjincludegraphics[width=\imgw,height=\imgh,cfbox=correct 2pt 0pt]{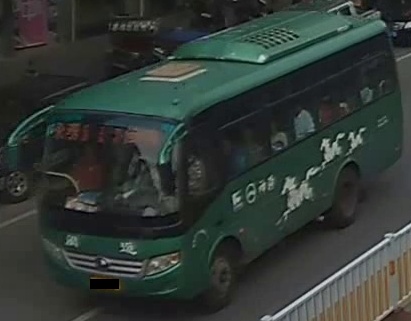}\\[-1pt]
    {\tiny\textbf{\textcolor{correct}{ID:~66}}}
  \end{tabular} &
    \begin{tabular}{@{}c@{}}
    \adjincludegraphics[width=\imgw,height=\imgh,cfbox=correct 2pt 0pt]{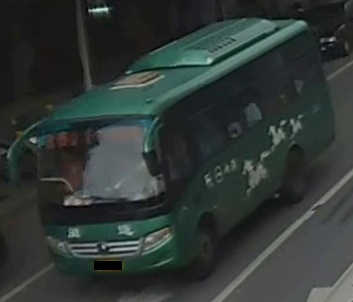}\\[-1pt]
    {\tiny\textbf{\textcolor{correct}{ID:~66}}}
  \end{tabular} &
    \begin{tabular}{@{}c@{}}
    \adjincludegraphics[width=\imgw,height=\imgh,cfbox=correct 2pt 0pt]{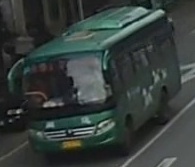}\\[-1pt]
    {\tiny\textbf{\textcolor{correct}{ID:~66}}}
  \end{tabular} &
    \begin{tabular}{@{}c@{}}
    \adjincludegraphics[width=\imgw,height=\imgh,cfbox=correct 2pt 0pt]{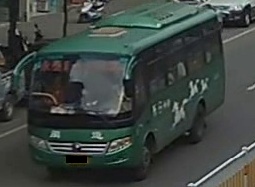}\\[-1pt]
    {\tiny\textbf{\textcolor{correct}{ID:~66}}}
  \end{tabular} &
    \begin{tabular}{@{}c@{}}
    \adjincludegraphics[width=\imgw,height=\imgh,cfbox=correct 2pt 0pt]{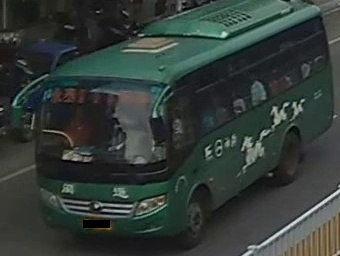}\\[-1pt]
    {\tiny\textbf{\textcolor{correct}{ID:~66}}}
  \end{tabular} &
    \begin{tabular}{@{}c@{}}
    \adjincludegraphics[width=\imgw,height=\imgh,cfbox=correct 2pt 0pt]{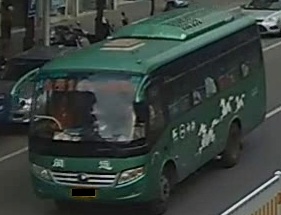}\\[-1pt]
    {\tiny\textbf{\textcolor{correct}{ID:~66}}}
  \end{tabular} &
    \begin{tabular}{@{}c@{}}
    \adjincludegraphics[width=\imgw,height=\imgh,cfbox=correct 2pt 0pt]{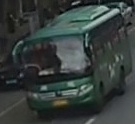}\\[-1pt]
    {\tiny\textbf{\textcolor{correct}{ID:~66}}}
  \end{tabular} &
    \begin{tabular}{@{}c@{}}
    \adjincludegraphics[width=\imgw,height=\imgh,cfbox=correct 2pt 0pt]{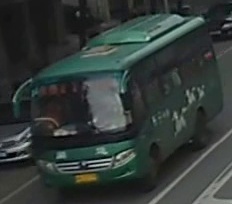}\\[-1pt]
    {\tiny\textbf{\textcolor{correct}{ID:~66}}}
  \end{tabular} \\[2pt]
  \hline \\[0.1pt]

  \rotatebox{90}{\parbox{20mm}{\centering\footnotesize\textbf{No BRM}}} &
    \begin{tabular}{@{}c@{}}
    \adjincludegraphics[width=\imgw,height=\imgh,cfbox=querycol 2pt 0pt]{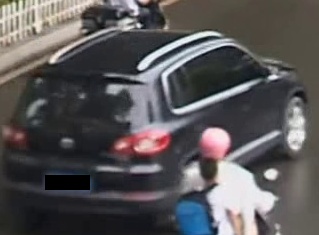}\\[-1pt]
    {\tiny\textbf{\textcolor{querycol}{ID:~615}}}
  \end{tabular} &
    \begin{tabular}{@{}c@{}}
    \adjincludegraphics[width=\imgw,height=\imgh,cfbox=wrong 2pt 0pt]{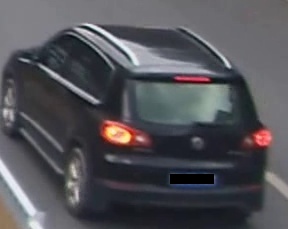}\\[-1pt]
    {\tiny\textbf{\textcolor{wrong}{ID:~585}}}
  \end{tabular} &
    \begin{tabular}{@{}c@{}}
    \adjincludegraphics[width=\imgw,height=\imgh,cfbox=wrong 2pt 0pt]{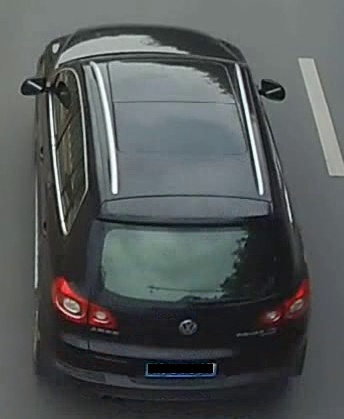}\\[-1pt]
    {\tiny\textbf{\textcolor{wrong}{ID:~585}}}
  \end{tabular} &
    \begin{tabular}{@{}c@{}}
    \adjincludegraphics[width=\imgw,height=\imgh,cfbox=wrong 2pt 0pt]{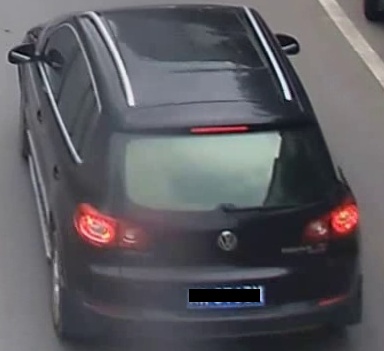}\\[-1pt]
    {\tiny\textbf{\textcolor{wrong}{ID:~585}}}
  \end{tabular} &
    \begin{tabular}{@{}c@{}}
    \adjincludegraphics[width=\imgw,height=\imgh,cfbox=correct 2pt 0pt]{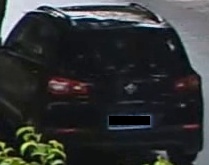}\\[-1pt]
    {\tiny\textbf{\textcolor{correct}{ID:~615}}}
  \end{tabular} &
    \begin{tabular}{@{}c@{}}
    \adjincludegraphics[width=\imgw,height=\imgh,cfbox=wrong 2pt 0pt]{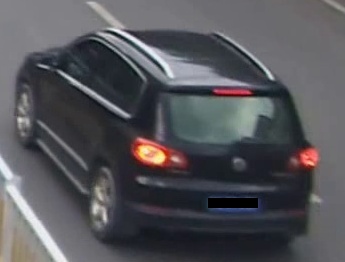}\\[-1pt]
    {\tiny\textbf{\textcolor{wrong}{ID:~585}}}
  \end{tabular} &
    \begin{tabular}{@{}c@{}}
    \adjincludegraphics[width=\imgw,height=\imgh,cfbox=correct 2pt 0pt]{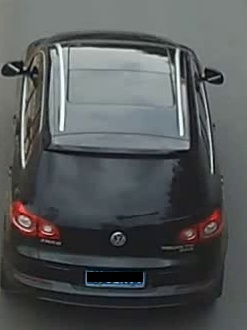}\\[-1pt]
    {\tiny\textbf{\textcolor{correct}{ID:~615}}}
  \end{tabular} &
    \begin{tabular}{@{}c@{}}
    \adjincludegraphics[width=\imgw,height=\imgh,cfbox=wrong 2pt 0pt]{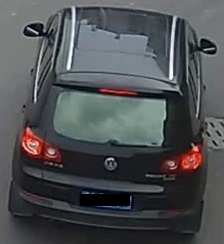}\\[-1pt]
    {\tiny\textbf{\textcolor{wrong}{ID:~585}}}
  \end{tabular} &
    \begin{tabular}{@{}c@{}}
    \adjincludegraphics[width=\imgw,height=\imgh,cfbox=wrong 2pt 0pt]{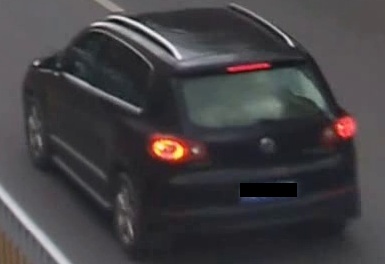}\\[-1pt]
    {\tiny\textbf{\textcolor{wrong}{ID:~585}}}
  \end{tabular} &
    \begin{tabular}{@{}c@{}}
    \adjincludegraphics[width=\imgw,height=\imgh,cfbox=wrong 2pt 0pt]{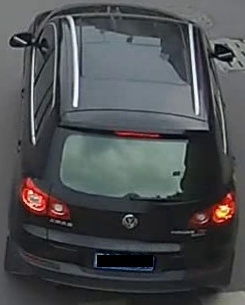}\\[-1pt]
    {\tiny\textbf{\textcolor{wrong}{ID:~585}}}
  \end{tabular} &
    \begin{tabular}{@{}c@{}}
    \adjincludegraphics[width=\imgw,height=\imgh,cfbox=wrong 2pt 0pt]{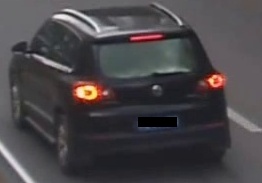}\\[-1pt]
    {\tiny\textbf{\textcolor{wrong}{ID:~585}}}
  \end{tabular} \\[2pt]
  \rotatebox{90}{\parbox{20mm}{\centering\footnotesize\textbf{BRM}}} &
    \begin{tabular}{@{}c@{}}
    \adjincludegraphics[width=\imgw,height=\imgh,cfbox=querycol 2pt 0pt]{supp/images/part3/q02_pid0615.jpg}\\[-1pt]
    {\tiny\textbf{\textcolor{querycol}{ID:~615}}}
  \end{tabular} &
    \begin{tabular}{@{}c@{}}
    \adjincludegraphics[width=\imgw,height=\imgh,cfbox=correct 2pt 0pt]{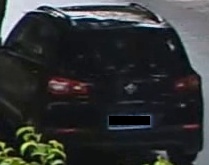}\\[-1pt]
    {\tiny\textbf{\textcolor{correct}{ID:~615}}}
  \end{tabular} &
    \begin{tabular}{@{}c@{}}
    \adjincludegraphics[width=\imgw,height=\imgh,cfbox=correct 2pt 0pt]{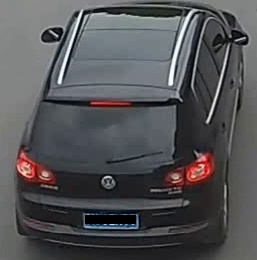}\\[-1pt]
    {\tiny\textbf{\textcolor{correct}{ID:~615}}}
  \end{tabular} &
    \begin{tabular}{@{}c@{}}
    \adjincludegraphics[width=\imgw,height=\imgh,cfbox=correct 2pt 0pt]{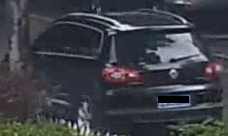}\\[-1pt]
    {\tiny\textbf{\textcolor{correct}{ID:~615}}}
  \end{tabular} &
    \begin{tabular}{@{}c@{}}
    \adjincludegraphics[width=\imgw,height=\imgh,cfbox=correct 2pt 0pt]{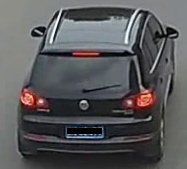}\\[-1pt]
    {\tiny\textbf{\textcolor{correct}{ID:~615}}}
  \end{tabular} &
    \begin{tabular}{@{}c@{}}
    \adjincludegraphics[width=\imgw,height=\imgh,cfbox=correct 2pt 0pt]{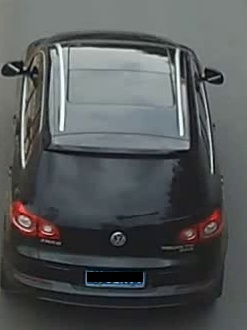}\\[-1pt]
    {\tiny\textbf{\textcolor{correct}{ID:~615}}}
  \end{tabular} &
    \begin{tabular}{@{}c@{}}
    \adjincludegraphics[width=\imgw,height=\imgh,cfbox=correct 2pt 0pt]{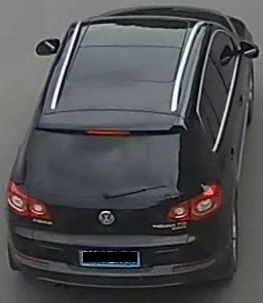}\\[-1pt]
    {\tiny\textbf{\textcolor{correct}{ID:~615}}}
  \end{tabular} &
    \begin{tabular}{@{}c@{}}
    \adjincludegraphics[width=\imgw,height=\imgh,cfbox=correct 2pt 0pt]{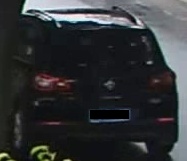}\\[-1pt]
    {\tiny\textbf{\textcolor{correct}{ID:~615}}}
  \end{tabular} &
    \begin{tabular}{@{}c@{}}
    \adjincludegraphics[width=\imgw,height=\imgh,cfbox=wrong 2pt 0pt]{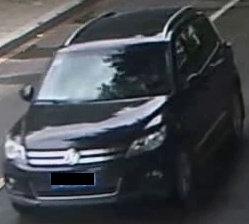}\\[-1pt]
    {\tiny\textbf{\textcolor{wrong}{ID:~585}}}
  \end{tabular} &
    \begin{tabular}{@{}c@{}}
    \adjincludegraphics[width=\imgw,height=\imgh,cfbox=correct 2pt 0pt]{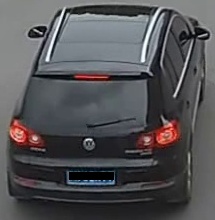}\\[-1pt]
    {\tiny\textbf{\textcolor{correct}{ID:~615}}}
  \end{tabular} &
    \begin{tabular}{@{}c@{}}
    \adjincludegraphics[width=\imgw,height=\imgh,cfbox=correct 2pt 0pt]{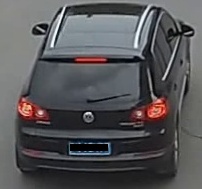}\\[-1pt]
    {\tiny\textbf{\textcolor{correct}{ID:~615}}}
  \end{tabular} \\[2pt]
  \hline \\[0.1pt]

  \rotatebox{90}{\parbox{20mm}{\centering\footnotesize\textbf{No BRM}}} &
    \begin{tabular}{@{}c@{}}
    \adjincludegraphics[width=\imgw,height=\imgh,cfbox=querycol 2pt 0pt]{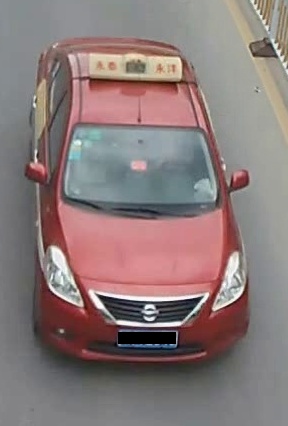}\\[-1pt]
    {\tiny\textbf{\textcolor{querycol}{ID:~38}}}
  \end{tabular} &
    \begin{tabular}{@{}c@{}}
    \adjincludegraphics[width=\imgw,height=\imgh,cfbox=correct 2pt 0pt]{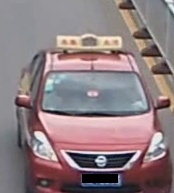}\\[-1pt]
    {\tiny\textbf{\textcolor{correct}{ID:~38}}}
  \end{tabular} &
    \begin{tabular}{@{}c@{}}
    \adjincludegraphics[width=\imgw,height=\imgh,cfbox=correct 2pt 0pt]{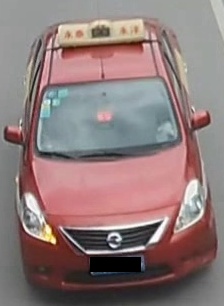}\\[-1pt]
    {\tiny\textbf{\textcolor{correct}{ID:~38}}}
  \end{tabular} &
    \begin{tabular}{@{}c@{}}
    \adjincludegraphics[width=\imgw,height=\imgh,cfbox=wrong 2pt 0pt]{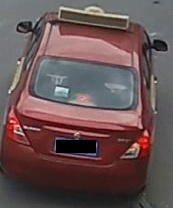}\\[-1pt]
    {\tiny\textbf{\textcolor{wrong}{ID:~35}}}
  \end{tabular} &
    \begin{tabular}{@{}c@{}}
    \adjincludegraphics[width=\imgw,height=\imgh,cfbox=correct 2pt 0pt]{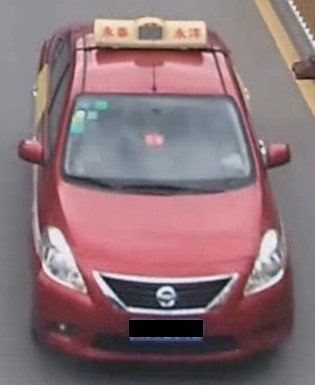}\\[-1pt]
    {\tiny\textbf{\textcolor{correct}{ID:~38}}}
  \end{tabular} &
    \begin{tabular}{@{}c@{}}
    \adjincludegraphics[width=\imgw,height=\imgh,cfbox=wrong 2pt 0pt]{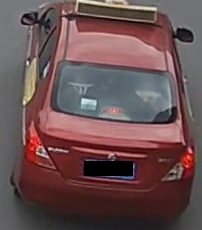}\\[-1pt]
    {\tiny\textbf{\textcolor{wrong}{ID:~35}}}
  \end{tabular} &
    \begin{tabular}{@{}c@{}}
    \adjincludegraphics[width=\imgw,height=\imgh,cfbox=wrong 2pt 0pt]{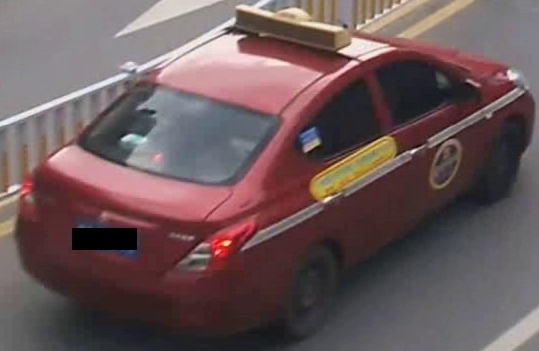}\\[-1pt]
    {\tiny\textbf{\textcolor{wrong}{ID:~35}}}
  \end{tabular} &
    \begin{tabular}{@{}c@{}}
    \adjincludegraphics[width=\imgw,height=\imgh,cfbox=wrong 2pt 0pt]{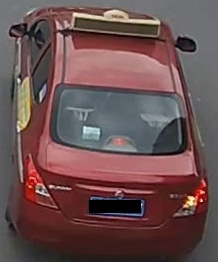}\\[-1pt]
    {\tiny\textbf{\textcolor{wrong}{ID:~35}}}
  \end{tabular} &
    \begin{tabular}{@{}c@{}}
    \adjincludegraphics[width=\imgw,height=\imgh,cfbox=wrong 2pt 0pt]{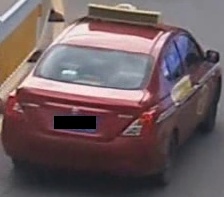}\\[-1pt]
    {\tiny\textbf{\textcolor{wrong}{ID:~35}}}
  \end{tabular} &
    \begin{tabular}{@{}c@{}}
    \adjincludegraphics[width=\imgw,height=\imgh,cfbox=correct 2pt 0pt]{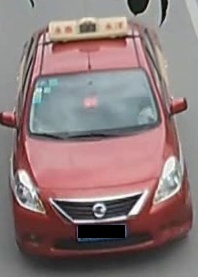}\\[-1pt]
    {\tiny\textbf{\textcolor{correct}{ID:~38}}}
  \end{tabular} &
    \begin{tabular}{@{}c@{}}
    \adjincludegraphics[width=\imgw,height=\imgh,cfbox=wrong 2pt 0pt]{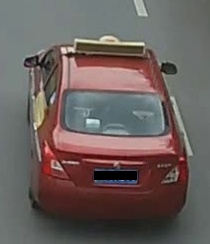}\\[-1pt]
    {\tiny\textbf{\textcolor{wrong}{ID:~35}}}
  \end{tabular} \\[2pt]
  \rotatebox{90}{\parbox{20mm}{\centering\footnotesize\textbf{BRM}}} &
    \begin{tabular}{@{}c@{}}
    \adjincludegraphics[width=\imgw,height=\imgh,cfbox=querycol 2pt 0pt]{supp/images/part3/q03_pid0038.jpg}\\[-1pt]
    {\tiny\textbf{\textcolor{querycol}{ID:~38}}}
  \end{tabular} &
    \begin{tabular}{@{}c@{}}
    \adjincludegraphics[width=\imgw,height=\imgh,cfbox=correct 2pt 0pt]{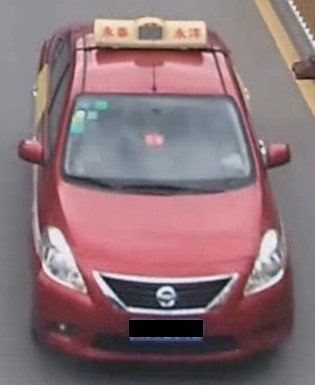}\\[-1pt]
    {\tiny\textbf{\textcolor{correct}{ID:~38}}}
  \end{tabular} &
    \begin{tabular}{@{}c@{}}
    \adjincludegraphics[width=\imgw,height=\imgh,cfbox=correct 2pt 0pt]{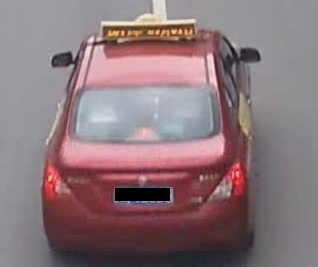}\\[-1pt]
    {\tiny\textbf{\textcolor{correct}{ID:~38}}}
  \end{tabular} &
    \begin{tabular}{@{}c@{}}
    \adjincludegraphics[width=\imgw,height=\imgh,cfbox=correct 2pt 0pt]{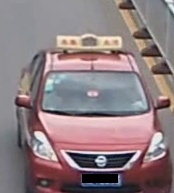}\\[-1pt]
    {\tiny\textbf{\textcolor{correct}{ID:~38}}}
  \end{tabular} &
    \begin{tabular}{@{}c@{}}
    \adjincludegraphics[width=\imgw,height=\imgh,cfbox=correct 2pt 0pt]{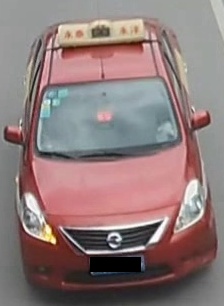}\\[-1pt]
    {\tiny\textbf{\textcolor{correct}{ID:~38}}}
  \end{tabular} &
    \begin{tabular}{@{}c@{}}
    \adjincludegraphics[width=\imgw,height=\imgh,cfbox=correct 2pt 0pt]{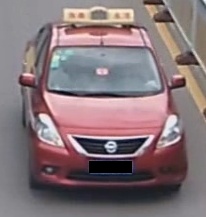}\\[-1pt]
    {\tiny\textbf{\textcolor{correct}{ID:~38}}}
  \end{tabular} &
    \begin{tabular}{@{}c@{}}
    \adjincludegraphics[width=\imgw,height=\imgh,cfbox=correct 2pt 0pt]{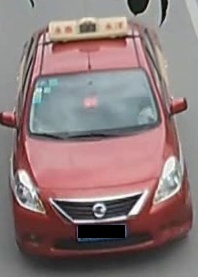}\\[-1pt]
    {\tiny\textbf{\textcolor{correct}{ID:~38}}}
  \end{tabular} &
    \begin{tabular}{@{}c@{}}
    \adjincludegraphics[width=\imgw,height=\imgh,cfbox=correct 2pt 0pt]{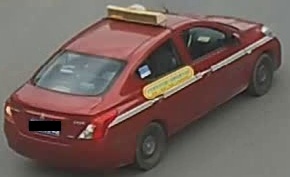}\\[-1pt]
    {\tiny\textbf{\textcolor{correct}{ID:~38}}}
  \end{tabular} &
    \begin{tabular}{@{}c@{}}
    \adjincludegraphics[width=\imgw,height=\imgh,cfbox=correct 2pt 0pt]{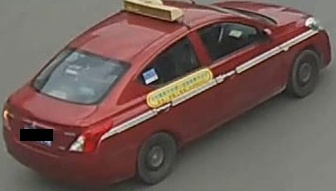}\\[-1pt]
    {\tiny\textbf{\textcolor{correct}{ID:~38}}}
  \end{tabular} &
    \begin{tabular}{@{}c@{}}
    \adjincludegraphics[width=\imgw,height=\imgh,cfbox=correct 2pt 0pt]{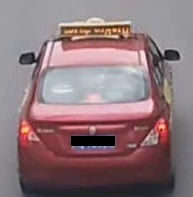}\\[-1pt]
    {\tiny\textbf{\textcolor{correct}{ID:~38}}}
  \end{tabular} &
    \begin{tabular}{@{}c@{}}
    \adjincludegraphics[width=\imgw,height=\imgh,cfbox=correct 2pt 0pt]{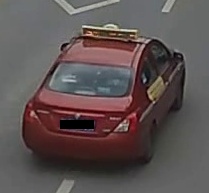}\\[-1pt]
    {\tiny\textbf{\textcolor{correct}{ID:~38}}}
  \end{tabular} \\[2pt]
  \end{tabular}}

}
  \caption{%
    Top-10 retrieval results for 3 VeRi-776 queries (PIDs 66, 615, 38), comparing the baseline without BRM against the BRM-trained model.
    \textcolor{correct}{Green} = correct match;\;
    \textcolor{wrong}{red} = incorrect match;\;
    \textcolor{querycol}{blue} = query image.}
  \label{fig:retrieval-page-01}
\end{figure*}

%% file: supp/msmt17_retrieval_comparison.tex
\begin{figure*}[p]
\setlength{\imgw}{20mm}\setlength{\imgh}{40mm}%
\resizebox{\textwidth}{!}{%
  \centering

  {\setlength{\tabcolsep}{2pt}%
  \begin{tabular}{m{6mm}ccccccccccc}
   & \textbf{Query} & \textcolor{milestone}{\textbf{R-1}} &  &  &  & \textcolor{milestone}{\textbf{R-5}} &  &  &  &  & \textcolor{milestone}{\textbf{R-10}} \\[2pt]
  \rotatebox{90}{\parbox{20mm}{\centering\footnotesize\textbf{No BRM}}} &
    \begin{tabular}{@{}c@{}}
    \adjincludegraphics[width=\imgw,height=\imgh,cfbox=querycol 2pt 0pt]{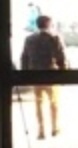}\\[-1pt]
    {\tiny\textbf{\textcolor{querycol}{ID:~702}}}
  \end{tabular} &
    \begin{tabular}{@{}c@{}}
    \adjincludegraphics[width=\imgw,height=\imgh,cfbox=wrong 2pt 0pt]{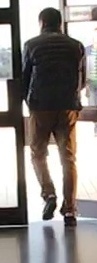}\\[-1pt]
    {\tiny\textbf{\textcolor{wrong}{ID:~2250}}}
  \end{tabular} &
    \begin{tabular}{@{}c@{}}
    \adjincludegraphics[width=\imgw,height=\imgh,cfbox=correct 2pt 0pt]{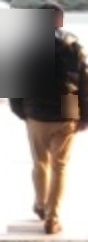}\\[-1pt]
    {\tiny\textbf{\textcolor{correct}{ID:~702}}}
  \end{tabular} &
    \begin{tabular}{@{}c@{}}
    \adjincludegraphics[width=\imgw,height=\imgh,cfbox=wrong 2pt 0pt]{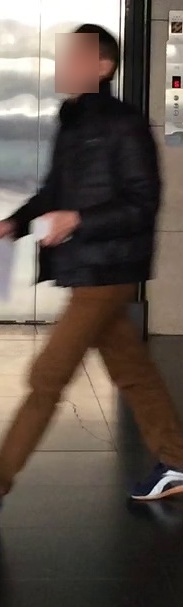}\\[-1pt]
    {\tiny\textbf{\textcolor{wrong}{ID:~2250}}}
  \end{tabular} &
    \begin{tabular}{@{}c@{}}
    \adjincludegraphics[width=\imgw,height=\imgh,cfbox=wrong 2pt 0pt]{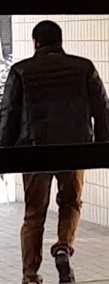}\\[-1pt]
    {\tiny\textbf{\textcolor{wrong}{ID:~2250}}}
  \end{tabular} &
    \begin{tabular}{@{}c@{}}
    \adjincludegraphics[width=\imgw,height=\imgh,cfbox=wrong 2pt 0pt]{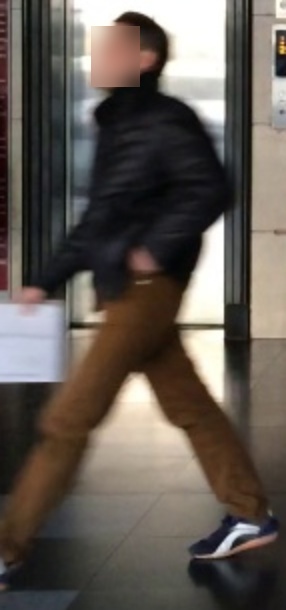}\\[-1pt]
    {\tiny\textbf{\textcolor{wrong}{ID:~2250}}}
  \end{tabular} &
    \begin{tabular}{@{}c@{}}
    \adjincludegraphics[width=\imgw,height=\imgh,cfbox=wrong 2pt 0pt]{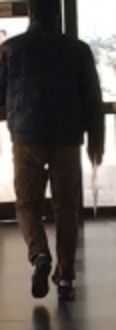}\\[-1pt]
    {\tiny\textbf{\textcolor{wrong}{ID:~2250}}}
  \end{tabular} &
    \begin{tabular}{@{}c@{}}
    \adjincludegraphics[width=\imgw,height=\imgh,cfbox=wrong 2pt 0pt]{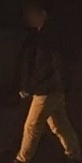}\\[-1pt]
    {\tiny\textbf{\textcolor{wrong}{ID:~691}}}
  \end{tabular} &
    \begin{tabular}{@{}c@{}}
    \adjincludegraphics[width=\imgw,height=\imgh,cfbox=wrong 2pt 0pt]{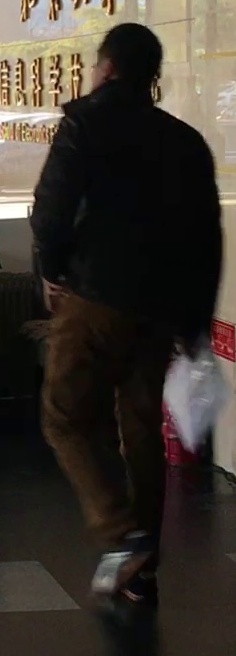}\\[-1pt]
    {\tiny\textbf{\textcolor{wrong}{ID:~2250}}}
  \end{tabular} &
    \begin{tabular}{@{}c@{}}
    \adjincludegraphics[width=\imgw,height=\imgh,cfbox=wrong 2pt 0pt]{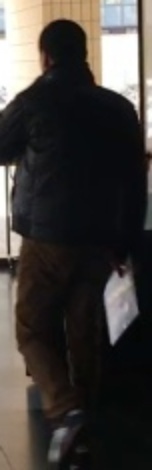}\\[-1pt]
    {\tiny\textbf{\textcolor{wrong}{ID:~2250}}}
  \end{tabular} &
    \begin{tabular}{@{}c@{}}
    \adjincludegraphics[width=\imgw,height=\imgh,cfbox=wrong 2pt 0pt]{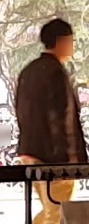}\\[-1pt]
    {\tiny\textbf{\textcolor{wrong}{ID:~691}}}
  \end{tabular} \\[2pt]
  \rotatebox{90}{\parbox{20mm}{\centering\footnotesize\textbf{BRM}}} &
    \begin{tabular}{@{}c@{}}
    \adjincludegraphics[width=\imgw,height=\imgh,cfbox=querycol 2pt 0pt]{supp/images_msmt17/part3/q01_pid0702.jpg}\\[-1pt]
    {\tiny\textbf{\textcolor{querycol}{ID:~702}}}
  \end{tabular} &
    \begin{tabular}{@{}c@{}}
    \adjincludegraphics[width=\imgw,height=\imgh,cfbox=correct 2pt 0pt]{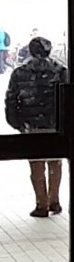}\\[-1pt]
    {\tiny\textbf{\textcolor{correct}{ID:~702}}}
  \end{tabular} &
    \begin{tabular}{@{}c@{}}
    \adjincludegraphics[width=\imgw,height=\imgh,cfbox=correct 2pt 0pt]{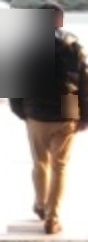}\\[-1pt]
    {\tiny\textbf{\textcolor{correct}{ID:~702}}}
  \end{tabular} &
    \begin{tabular}{@{}c@{}}
    \adjincludegraphics[width=\imgw,height=\imgh,cfbox=correct 2pt 0pt]{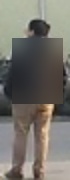}\\[-1pt]
    {\tiny\textbf{\textcolor{correct}{ID:~702}}}
  \end{tabular} &
    \begin{tabular}{@{}c@{}}
    \adjincludegraphics[width=\imgw,height=\imgh,cfbox=correct 2pt 0pt]{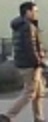}\\[-1pt]
    {\tiny\textbf{\textcolor{correct}{ID:~702}}}
  \end{tabular} &
    \begin{tabular}{@{}c@{}}
    \adjincludegraphics[width=\imgw,height=\imgh,cfbox=correct 2pt 0pt]{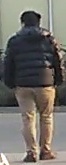}\\[-1pt]
    {\tiny\textbf{\textcolor{correct}{ID:~702}}}
  \end{tabular} &
    \begin{tabular}{@{}c@{}}
    \adjincludegraphics[width=\imgw,height=\imgh,cfbox=correct 2pt 0pt]{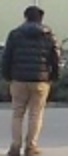}\\[-1pt]
    {\tiny\textbf{\textcolor{correct}{ID:~702}}}
  \end{tabular} &
    \begin{tabular}{@{}c@{}}
    \adjincludegraphics[width=\imgw,height=\imgh,cfbox=correct 2pt 0pt]{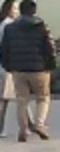}\\[-1pt]
    {\tiny\textbf{\textcolor{correct}{ID:~702}}}
  \end{tabular} &
    \begin{tabular}{@{}c@{}}
    \adjincludegraphics[width=\imgw,height=\imgh,cfbox=correct 2pt 0pt]{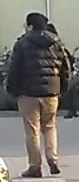}\\[-1pt]
    {\tiny\textbf{\textcolor{correct}{ID:~702}}}
  \end{tabular} &
    \begin{tabular}{@{}c@{}}
    \adjincludegraphics[width=\imgw,height=\imgh,cfbox=correct 2pt 0pt]{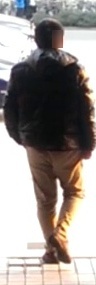}\\[-1pt]
    {\tiny\textbf{\textcolor{correct}{ID:~702}}}
  \end{tabular} &
    \begin{tabular}{@{}c@{}}
    \adjincludegraphics[width=\imgw,height=\imgh,cfbox=correct 2pt 0pt]{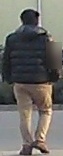}\\[-1pt]
    {\tiny\textbf{\textcolor{correct}{ID:~702}}}
  \end{tabular} \\[2pt]
  \hline \\[0.1pt]

  \rotatebox{90}{\parbox{20mm}{\centering\footnotesize\textbf{No BRM}}} &
    \begin{tabular}{@{}c@{}}
    \adjincludegraphics[width=\imgw,height=\imgh,cfbox=querycol 2pt 0pt]{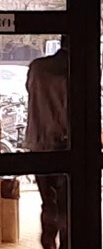}\\[-1pt]
    {\tiny\textbf{\textcolor{querycol}{ID:~803}}}
  \end{tabular} &
    \begin{tabular}{@{}c@{}}
    \adjincludegraphics[width=\imgw,height=\imgh,cfbox=wrong 2pt 0pt]{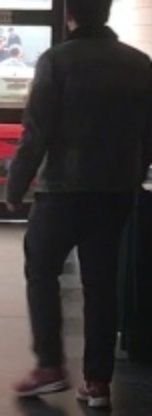}\\[-1pt]
    {\tiny\textbf{\textcolor{wrong}{ID:~396}}}
  \end{tabular} &
    \begin{tabular}{@{}c@{}}
    \adjincludegraphics[width=\imgw,height=\imgh,cfbox=wrong 2pt 0pt]{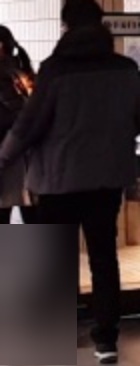}\\[-1pt]
    {\tiny\textbf{\textcolor{wrong}{ID:~820}}}
  \end{tabular} &
    \begin{tabular}{@{}c@{}}
    \adjincludegraphics[width=\imgw,height=\imgh,cfbox=wrong 2pt 0pt]{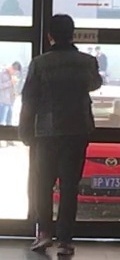}\\[-1pt]
    {\tiny\textbf{\textcolor{wrong}{ID:~396}}}
  \end{tabular} &
    \begin{tabular}{@{}c@{}}
    \adjincludegraphics[width=\imgw,height=\imgh,cfbox=wrong 2pt 0pt]{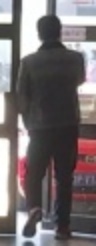}\\[-1pt]
    {\tiny\textbf{\textcolor{wrong}{ID:~396}}}
  \end{tabular} &
    \begin{tabular}{@{}c@{}}
    \adjincludegraphics[width=\imgw,height=\imgh,cfbox=wrong 2pt 0pt]{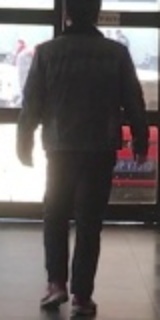}\\[-1pt]
    {\tiny\textbf{\textcolor{wrong}{ID:~396}}}
  \end{tabular} &
    \begin{tabular}{@{}c@{}}
    \adjincludegraphics[width=\imgw,height=\imgh,cfbox=wrong 2pt 0pt]{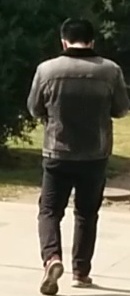}\\[-1pt]
    {\tiny\textbf{\textcolor{wrong}{ID:~396}}}
  \end{tabular} &
    \begin{tabular}{@{}c@{}}
    \adjincludegraphics[width=\imgw,height=\imgh,cfbox=wrong 2pt 0pt]{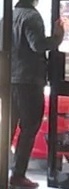}\\[-1pt]
    {\tiny\textbf{\textcolor{wrong}{ID:~396}}}
  \end{tabular} &
    \begin{tabular}{@{}c@{}}
    \adjincludegraphics[width=\imgw,height=\imgh,cfbox=wrong 2pt 0pt]{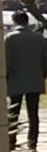}\\[-1pt]
    {\tiny\textbf{\textcolor{wrong}{ID:~1931}}}
  \end{tabular} &
    \begin{tabular}{@{}c@{}}
    \adjincludegraphics[width=\imgw,height=\imgh,cfbox=wrong 2pt 0pt]{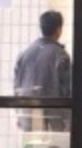}\\[-1pt]
    {\tiny\textbf{\textcolor{wrong}{ID:~1862}}}
  \end{tabular} &
    \begin{tabular}{@{}c@{}}
    \adjincludegraphics[width=\imgw,height=\imgh,cfbox=wrong 2pt 0pt]{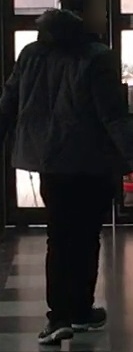}\\[-1pt]
    {\tiny\textbf{\textcolor{wrong}{ID:~820}}}
  \end{tabular} \\[2pt]
  \rotatebox{90}{\parbox{20mm}{\centering\footnotesize\textbf{BRM}}} &
    \begin{tabular}{@{}c@{}}
    \adjincludegraphics[width=\imgw,height=\imgh,cfbox=querycol 2pt 0pt]{supp/images_msmt17/part3/q02_pid0803.jpg}\\[-1pt]
    {\tiny\textbf{\textcolor{querycol}{ID:~803}}}
  \end{tabular} &
    \begin{tabular}{@{}c@{}}
    \adjincludegraphics[width=\imgw,height=\imgh,cfbox=correct 2pt 0pt]{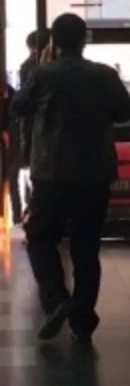}\\[-1pt]
    {\tiny\textbf{\textcolor{correct}{ID:~803}}}
  \end{tabular} &
    \begin{tabular}{@{}c@{}}
    \adjincludegraphics[width=\imgw,height=\imgh,cfbox=correct 2pt 0pt]{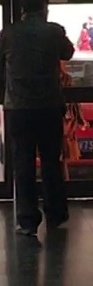}\\[-1pt]
    {\tiny\textbf{\textcolor{correct}{ID:~803}}}
  \end{tabular} &
    \begin{tabular}{@{}c@{}}
    \adjincludegraphics[width=\imgw,height=\imgh,cfbox=correct 2pt 0pt]{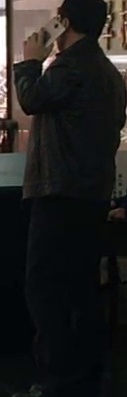}\\[-1pt]
    {\tiny\textbf{\textcolor{correct}{ID:~803}}}
  \end{tabular} &
    \begin{tabular}{@{}c@{}}
    \adjincludegraphics[width=\imgw,height=\imgh,cfbox=correct 2pt 0pt]{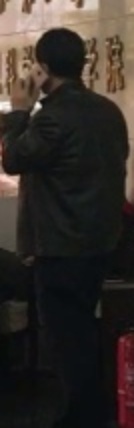}\\[-1pt]
    {\tiny\textbf{\textcolor{correct}{ID:~803}}}
  \end{tabular} &
    \begin{tabular}{@{}c@{}}
    \adjincludegraphics[width=\imgw,height=\imgh,cfbox=correct 2pt 0pt]{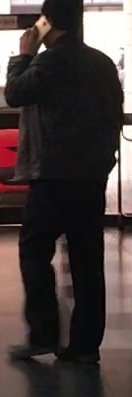}\\[-1pt]
    {\tiny\textbf{\textcolor{correct}{ID:~803}}}
  \end{tabular} &
    \begin{tabular}{@{}c@{}}
    \adjincludegraphics[width=\imgw,height=\imgh,cfbox=correct 2pt 0pt]{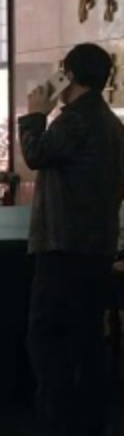}\\[-1pt]
    {\tiny\textbf{\textcolor{correct}{ID:~803}}}
  \end{tabular} &
    \begin{tabular}{@{}c@{}}
    \adjincludegraphics[width=\imgw,height=\imgh,cfbox=correct 2pt 0pt]{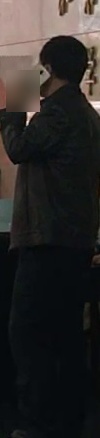}\\[-1pt]
    {\tiny\textbf{\textcolor{correct}{ID:~803}}}
  \end{tabular} &
    \begin{tabular}{@{}c@{}}
    \adjincludegraphics[width=\imgw,height=\imgh,cfbox=wrong 2pt 0pt]{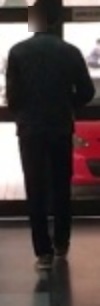}\\[-1pt]
    {\tiny\textbf{\textcolor{wrong}{ID:~831}}}
  \end{tabular} &
    \begin{tabular}{@{}c@{}}
    \adjincludegraphics[width=\imgw,height=\imgh,cfbox=correct 2pt 0pt]{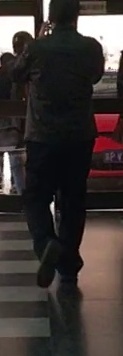}\\[-1pt]
    {\tiny\textbf{\textcolor{correct}{ID:~803}}}
  \end{tabular} &
    \begin{tabular}{@{}c@{}}
    \adjincludegraphics[width=\imgw,height=\imgh,cfbox=correct 2pt 0pt]{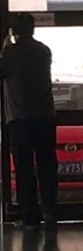}\\[-1pt]
    {\tiny\textbf{\textcolor{correct}{ID:~803}}}
  \end{tabular} \\[2pt]
  \hline \\[0.1pt]

  \rotatebox{90}{\parbox{20mm}{\centering\footnotesize\textbf{No BRM}}} &
    \begin{tabular}{@{}c@{}}
    \adjincludegraphics[width=\imgw,height=\imgh,cfbox=querycol 2pt 0pt]{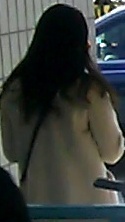}\\[-1pt]
    {\tiny\textbf{\textcolor{querycol}{ID:~1874}}}
  \end{tabular} &
    \begin{tabular}{@{}c@{}}
    \adjincludegraphics[width=\imgw,height=\imgh,cfbox=wrong 2pt 0pt]{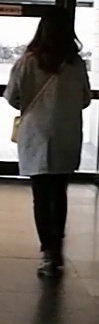}\\[-1pt]
    {\tiny\textbf{\textcolor{wrong}{ID:~1624}}}
  \end{tabular} &
    \begin{tabular}{@{}c@{}}
    \adjincludegraphics[width=\imgw,height=\imgh,cfbox=correct 2pt 0pt]{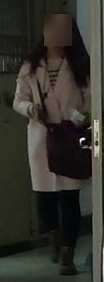}\\[-1pt]
    {\tiny\textbf{\textcolor{correct}{ID:~1874}}}
  \end{tabular} &
    \begin{tabular}{@{}c@{}}
    \adjincludegraphics[width=\imgw,height=\imgh,cfbox=wrong 2pt 0pt]{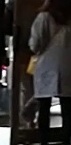}\\[-1pt]
    {\tiny\textbf{\textcolor{wrong}{ID:~1624}}}
  \end{tabular} &
    \begin{tabular}{@{}c@{}}
    \adjincludegraphics[width=\imgw,height=\imgh,cfbox=wrong 2pt 0pt]{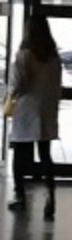}\\[-1pt]
    {\tiny\textbf{\textcolor{wrong}{ID:~1624}}}
  \end{tabular} &
    \begin{tabular}{@{}c@{}}
    \adjincludegraphics[width=\imgw,height=\imgh,cfbox=wrong 2pt 0pt]{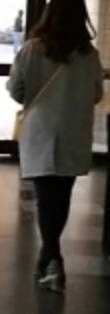}\\[-1pt]
    {\tiny\textbf{\textcolor{wrong}{ID:~1624}}}
  \end{tabular} &
    \begin{tabular}{@{}c@{}}
    \adjincludegraphics[width=\imgw,height=\imgh,cfbox=wrong 2pt 0pt]{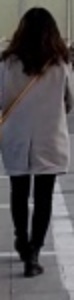}\\[-1pt]
    {\tiny\textbf{\textcolor{wrong}{ID:~1624}}}
  \end{tabular} &
    \begin{tabular}{@{}c@{}}
    \adjincludegraphics[width=\imgw,height=\imgh,cfbox=wrong 2pt 0pt]{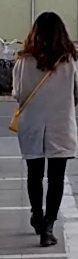}\\[-1pt]
    {\tiny\textbf{\textcolor{wrong}{ID:~1624}}}
  \end{tabular} &
    \begin{tabular}{@{}c@{}}
    \adjincludegraphics[width=\imgw,height=\imgh,cfbox=wrong 2pt 0pt]{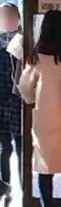}\\[-1pt]
    {\tiny\textbf{\textcolor{wrong}{ID:~1860}}}
  \end{tabular} &
    \begin{tabular}{@{}c@{}}
    \adjincludegraphics[width=\imgw,height=\imgh,cfbox=wrong 2pt 0pt]{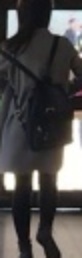}\\[-1pt]
    {\tiny\textbf{\textcolor{wrong}{ID:~1897}}}
  \end{tabular} &
    \begin{tabular}{@{}c@{}}
    \adjincludegraphics[width=\imgw,height=\imgh,cfbox=wrong 2pt 0pt]{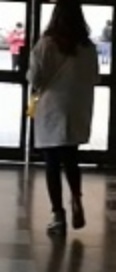}\\[-1pt]
    {\tiny\textbf{\textcolor{wrong}{ID:~1624}}}
  \end{tabular} \\[2pt]
  \rotatebox{90}{\parbox{20mm}{\centering\footnotesize\textbf{BRM}}} &
    \begin{tabular}{@{}c@{}}
    \adjincludegraphics[width=\imgw,height=\imgh,cfbox=querycol 2pt 0pt]{supp/images_msmt17/part3/q03_pid1874.jpg}\\[-1pt]
    {\tiny\textbf{\textcolor{querycol}{ID:~1874}}}
  \end{tabular} &
    \begin{tabular}{@{}c@{}}
    \adjincludegraphics[width=\imgw,height=\imgh,cfbox=correct 2pt 0pt]{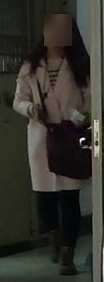}\\[-1pt]
    {\tiny\textbf{\textcolor{correct}{ID:~1874}}}
  \end{tabular} &
    \begin{tabular}{@{}c@{}}
    \adjincludegraphics[width=\imgw,height=\imgh,cfbox=correct 2pt 0pt]{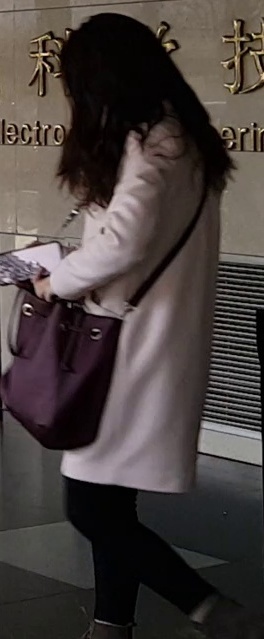}\\[-1pt]
    {\tiny\textbf{\textcolor{correct}{ID:~1874}}}
  \end{tabular} &
    \begin{tabular}{@{}c@{}}
    \adjincludegraphics[width=\imgw,height=\imgh,cfbox=correct 2pt 0pt]{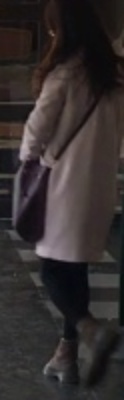}\\[-1pt]
    {\tiny\textbf{\textcolor{correct}{ID:~1874}}}
  \end{tabular} &
    \begin{tabular}{@{}c@{}}
    \adjincludegraphics[width=\imgw,height=\imgh,cfbox=correct 2pt 0pt]{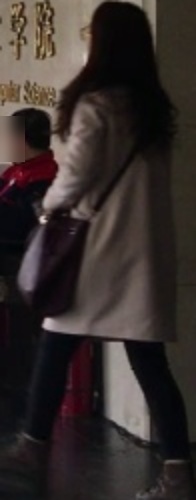}\\[-1pt]
    {\tiny\textbf{\textcolor{correct}{ID:~1874}}}
  \end{tabular} &
    \begin{tabular}{@{}c@{}}
    \adjincludegraphics[width=\imgw,height=\imgh,cfbox=correct 2pt 0pt]{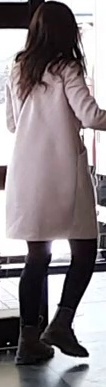}\\[-1pt]
    {\tiny\textbf{\textcolor{correct}{ID:~1874}}}
  \end{tabular} &
    \begin{tabular}{@{}c@{}}
    \adjincludegraphics[width=\imgw,height=\imgh,cfbox=correct 2pt 0pt]{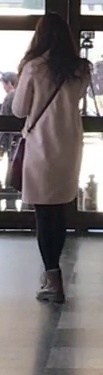}\\[-1pt]
    {\tiny\textbf{\textcolor{correct}{ID:~1874}}}
  \end{tabular} &
    \begin{tabular}{@{}c@{}}
    \adjincludegraphics[width=\imgw,height=\imgh,cfbox=correct 2pt 0pt]{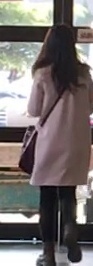}\\[-1pt]
    {\tiny\textbf{\textcolor{correct}{ID:~1874}}}
  \end{tabular} &
    \begin{tabular}{@{}c@{}}
    \adjincludegraphics[width=\imgw,height=\imgh,cfbox=correct 2pt 0pt]{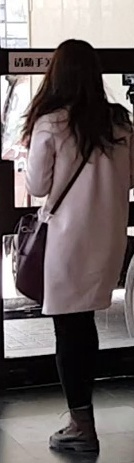}\\[-1pt]
    {\tiny\textbf{\textcolor{correct}{ID:~1874}}}
  \end{tabular} &
    \begin{tabular}{@{}c@{}}
    \adjincludegraphics[width=\imgw,height=\imgh,cfbox=correct 2pt 0pt]{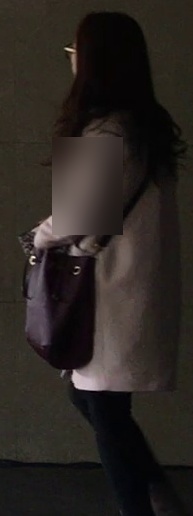}\\[-1pt]
    {\tiny\textbf{\textcolor{correct}{ID:~1874}}}
  \end{tabular} &
    \begin{tabular}{@{}c@{}}
    \adjincludegraphics[width=\imgw,height=\imgh,cfbox=correct 2pt 0pt]{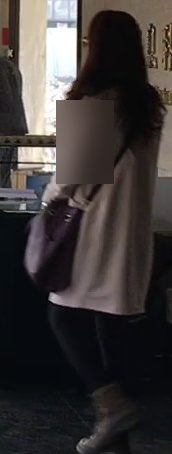}\\[-1pt]
    {\tiny\textbf{\textcolor{correct}{ID:~1874}}}
  \end{tabular} \\[2pt]
  \end{tabular}}

}
  \caption{%
    Top-10 retrieval results for 3 MSMT17 queries (PIDs 702, 803, 1874), comparing the baseline without BRM against the BRM-trained model.
    \textcolor{correct}{Green} = correct match;\;
    \textcolor{wrong}{red} = incorrect match;\;
    \textcolor{querycol}{blue} = query image.}
  \label{fig:retrieval-msmt17-01}
\end{figure*}

\begin{figure*}[p]
\setlength{\imgw}{20mm}\setlength{\imgh}{40mm}%
\resizebox{\textwidth}{!}{%
  \centering

  {\setlength{\tabcolsep}{2pt}%
  \begin{tabular}{m{6mm}ccccccccccc}
   & \textbf{Query} & \textcolor{milestone}{\textbf{R-1}} &  &  &  & \textcolor{milestone}{\textbf{R-5}} &  &  &  &  & \textcolor{milestone}{\textbf{R-10}} \\[2pt]
  \rotatebox{90}{\parbox{20mm}{\centering\footnotesize\textbf{No BRM}}} &
    \begin{tabular}{@{}c@{}}
    \adjincludegraphics[width=\imgw,height=\imgh,cfbox=querycol 2pt 0pt]{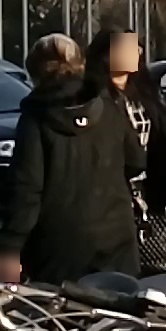}\\[-1pt]
    {\tiny\textbf{\textcolor{querycol}{ID:~2046}}}
  \end{tabular} &
    \begin{tabular}{@{}c@{}}
    \adjincludegraphics[width=\imgw,height=\imgh,cfbox=wrong 2pt 0pt]{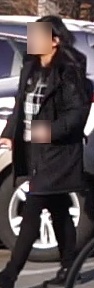}\\[-1pt]
    {\tiny\textbf{\textcolor{wrong}{ID:~2216}}}
  \end{tabular} &
    \begin{tabular}{@{}c@{}}
    \adjincludegraphics[width=\imgw,height=\imgh,cfbox=wrong 2pt 0pt]{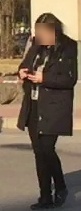}\\[-1pt]
    {\tiny\textbf{\textcolor{wrong}{ID:~2216}}}
  \end{tabular} &
    \begin{tabular}{@{}c@{}}
    \adjincludegraphics[width=\imgw,height=\imgh,cfbox=wrong 2pt 0pt]{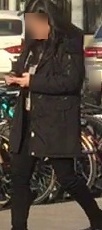}\\[-1pt]
    {\tiny\textbf{\textcolor{wrong}{ID:~2216}}}
  \end{tabular} &
    \begin{tabular}{@{}c@{}}
    \adjincludegraphics[width=\imgw,height=\imgh,cfbox=wrong 2pt 0pt]{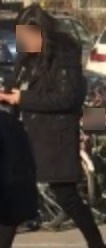}\\[-1pt]
    {\tiny\textbf{\textcolor{wrong}{ID:~2216}}}
  \end{tabular} &
    \begin{tabular}{@{}c@{}}
    \adjincludegraphics[width=\imgw,height=\imgh,cfbox=wrong 2pt 0pt]{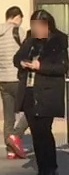}\\[-1pt]
    {\tiny\textbf{\textcolor{wrong}{ID:~2216}}}
  \end{tabular} &
    \begin{tabular}{@{}c@{}}
    \adjincludegraphics[width=\imgw,height=\imgh,cfbox=wrong 2pt 0pt]{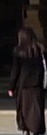}\\[-1pt]
    {\tiny\textbf{\textcolor{wrong}{ID:~2073}}}
  \end{tabular} &
    \begin{tabular}{@{}c@{}}
    \adjincludegraphics[width=\imgw,height=\imgh,cfbox=wrong 2pt 0pt]{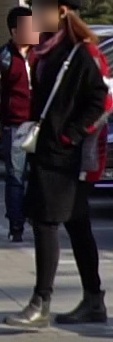}\\[-1pt]
    {\tiny\textbf{\textcolor{wrong}{ID:~61}}}
  \end{tabular} &
    \begin{tabular}{@{}c@{}}
    \adjincludegraphics[width=\imgw,height=\imgh,cfbox=wrong 2pt 0pt]{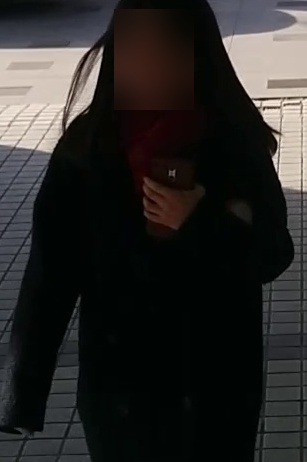}\\[-1pt]
    {\tiny\textbf{\textcolor{wrong}{ID:~1834}}}
  \end{tabular} &
    \begin{tabular}{@{}c@{}}
    \adjincludegraphics[width=\imgw,height=\imgh,cfbox=wrong 2pt 0pt]{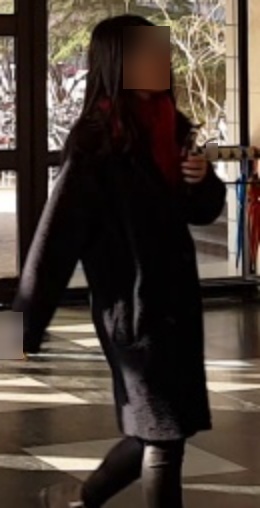}\\[-1pt]
    {\tiny\textbf{\textcolor{wrong}{ID:~1834}}}
  \end{tabular} &
    \begin{tabular}{@{}c@{}}
    \adjincludegraphics[width=\imgw,height=\imgh,cfbox=wrong 2pt 0pt]{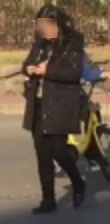}\\[-1pt]
    {\tiny\textbf{\textcolor{wrong}{ID:~2216}}}
  \end{tabular} \\[2pt]
  \rotatebox{90}{\parbox{20mm}{\centering\footnotesize\textbf{BRM}}} &
    \begin{tabular}{@{}c@{}}
    \adjincludegraphics[width=\imgw,height=\imgh,cfbox=querycol 2pt 0pt]{supp/images_msmt17/part3/q04_pid2046.jpg}\\[-1pt]
    {\tiny\textbf{\textcolor{querycol}{ID:~2046}}}
  \end{tabular} &
    \begin{tabular}{@{}c@{}}
    \adjincludegraphics[width=\imgw,height=\imgh,cfbox=correct 2pt 0pt]{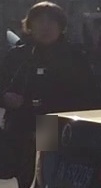}\\[-1pt]
    {\tiny\textbf{\textcolor{correct}{ID:~2046}}}
  \end{tabular} &
    \begin{tabular}{@{}c@{}}
    \adjincludegraphics[width=\imgw,height=\imgh,cfbox=correct 2pt 0pt]{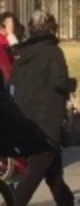}\\[-1pt]
    {\tiny\textbf{\textcolor{correct}{ID:~2046}}}
  \end{tabular} &
    \begin{tabular}{@{}c@{}}
    \adjincludegraphics[width=\imgw,height=\imgh,cfbox=correct 2pt 0pt]{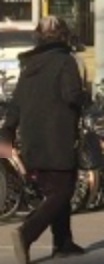}\\[-1pt]
    {\tiny\textbf{\textcolor{correct}{ID:~2046}}}
  \end{tabular} &
    \begin{tabular}{@{}c@{}}
    \adjincludegraphics[width=\imgw,height=\imgh,cfbox=correct 2pt 0pt]{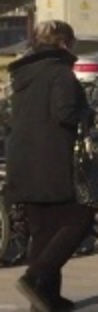}\\[-1pt]
    {\tiny\textbf{\textcolor{correct}{ID:~2046}}}
  \end{tabular} &
    \begin{tabular}{@{}c@{}}
    \adjincludegraphics[width=\imgw,height=\imgh,cfbox=correct 2pt 0pt]{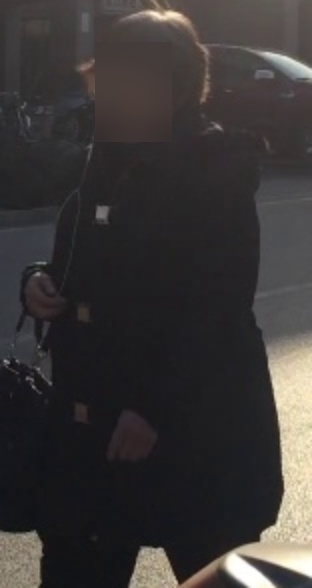}\\[-1pt]
    {\tiny\textbf{\textcolor{correct}{ID:~2046}}}
  \end{tabular} &
    \begin{tabular}{@{}c@{}}
    \adjincludegraphics[width=\imgw,height=\imgh,cfbox=correct 2pt 0pt]{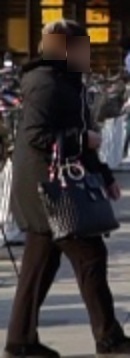}\\[-1pt]
    {\tiny\textbf{\textcolor{correct}{ID:~2046}}}
  \end{tabular} &
    \begin{tabular}{@{}c@{}}
    \adjincludegraphics[width=\imgw,height=\imgh,cfbox=wrong 2pt 0pt]{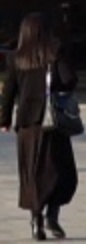}\\[-1pt]
    {\tiny\textbf{\textcolor{wrong}{ID:~2073}}}
  \end{tabular} &
    \begin{tabular}{@{}c@{}}
    \adjincludegraphics[width=\imgw,height=\imgh,cfbox=wrong 2pt 0pt]{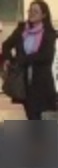}\\[-1pt]
    {\tiny\textbf{\textcolor{wrong}{ID:~2229}}}
  \end{tabular} &
    \begin{tabular}{@{}c@{}}
    \adjincludegraphics[width=\imgw,height=\imgh,cfbox=correct 2pt 0pt]{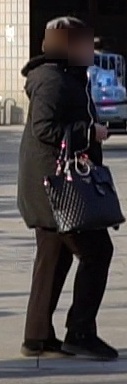}\\[-1pt]
    {\tiny\textbf{\textcolor{correct}{ID:~2046}}}
  \end{tabular} &
    \begin{tabular}{@{}c@{}}
    \adjincludegraphics[width=\imgw,height=\imgh,cfbox=correct 2pt 0pt]{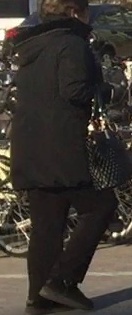}\\[-1pt]
    {\tiny\textbf{\textcolor{correct}{ID:~2046}}}
  \end{tabular} \\[2pt]
  \hline \\[0.1pt]

  \rotatebox{90}{\parbox{20mm}{\centering\footnotesize\textbf{No BRM}}} &
    \begin{tabular}{@{}c@{}}
    \adjincludegraphics[width=\imgw,height=\imgh,cfbox=querycol 2pt 0pt]{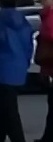}\\[-1pt]
    {\tiny\textbf{\textcolor{querycol}{ID:~2055}}}
  \end{tabular} &
    \begin{tabular}{@{}c@{}}
    \adjincludegraphics[width=\imgw,height=\imgh,cfbox=wrong 2pt 0pt]{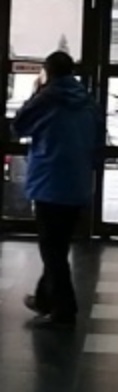}\\[-1pt]
    {\tiny\textbf{\textcolor{wrong}{ID:~1641}}}
  \end{tabular} &
    \begin{tabular}{@{}c@{}}
    \adjincludegraphics[width=\imgw,height=\imgh,cfbox=wrong 2pt 0pt]{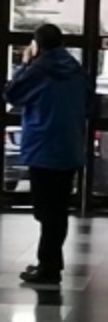}\\[-1pt]
    {\tiny\textbf{\textcolor{wrong}{ID:~1641}}}
  \end{tabular} &
    \begin{tabular}{@{}c@{}}
    \adjincludegraphics[width=\imgw,height=\imgh,cfbox=wrong 2pt 0pt]{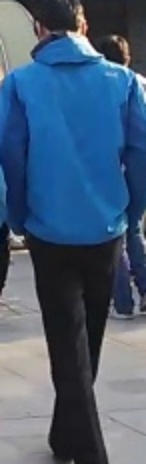}\\[-1pt]
    {\tiny\textbf{\textcolor{wrong}{ID:~2287}}}
  \end{tabular} &
    \begin{tabular}{@{}c@{}}
    \adjincludegraphics[width=\imgw,height=\imgh,cfbox=wrong 2pt 0pt]{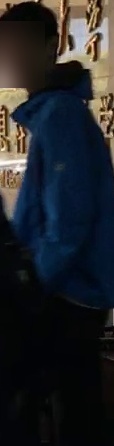}\\[-1pt]
    {\tiny\textbf{\textcolor{wrong}{ID:~2287}}}
  \end{tabular} &
    \begin{tabular}{@{}c@{}}
    \adjincludegraphics[width=\imgw,height=\imgh,cfbox=wrong 2pt 0pt]{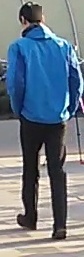}\\[-1pt]
    {\tiny\textbf{\textcolor{wrong}{ID:~2287}}}
  \end{tabular} &
    \begin{tabular}{@{}c@{}}
    \adjincludegraphics[width=\imgw,height=\imgh,cfbox=correct 2pt 0pt]{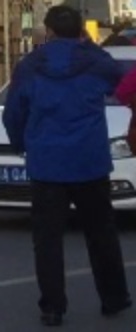}\\[-1pt]
    {\tiny\textbf{\textcolor{correct}{ID:~2055}}}
  \end{tabular} &
    \begin{tabular}{@{}c@{}}
    \adjincludegraphics[width=\imgw,height=\imgh,cfbox=correct 2pt 0pt]{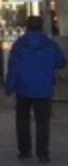}\\[-1pt]
    {\tiny\textbf{\textcolor{correct}{ID:~2055}}}
  \end{tabular} &
    \begin{tabular}{@{}c@{}}
    \adjincludegraphics[width=\imgw,height=\imgh,cfbox=wrong 2pt 0pt]{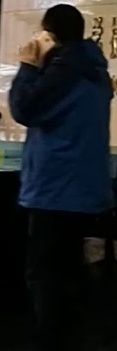}\\[-1pt]
    {\tiny\textbf{\textcolor{wrong}{ID:~1641}}}
  \end{tabular} &
    \begin{tabular}{@{}c@{}}
    \adjincludegraphics[width=\imgw,height=\imgh,cfbox=wrong 2pt 0pt]{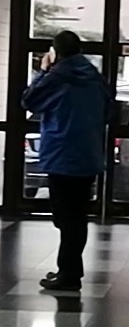}\\[-1pt]
    {\tiny\textbf{\textcolor{wrong}{ID:~1641}}}
  \end{tabular} &
    \begin{tabular}{@{}c@{}}
    \adjincludegraphics[width=\imgw,height=\imgh,cfbox=wrong 2pt 0pt]{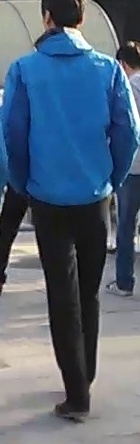}\\[-1pt]
    {\tiny\textbf{\textcolor{wrong}{ID:~2287}}}
  \end{tabular} \\[2pt]
  \rotatebox{90}{\parbox{20mm}{\centering\footnotesize\textbf{BRM}}} &
    \begin{tabular}{@{}c@{}}
    \adjincludegraphics[width=\imgw,height=\imgh,cfbox=querycol 2pt 0pt]{supp/images_msmt17/part3/q05_pid2055.jpg}\\[-1pt]
    {\tiny\textbf{\textcolor{querycol}{ID:~2055}}}
  \end{tabular} &
    \begin{tabular}{@{}c@{}}
    \adjincludegraphics[width=\imgw,height=\imgh,cfbox=correct 2pt 0pt]{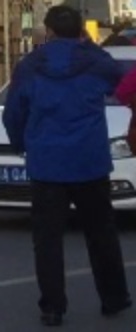}\\[-1pt]
    {\tiny\textbf{\textcolor{correct}{ID:~2055}}}
  \end{tabular} &
    \begin{tabular}{@{}c@{}}
    \adjincludegraphics[width=\imgw,height=\imgh,cfbox=correct 2pt 0pt]{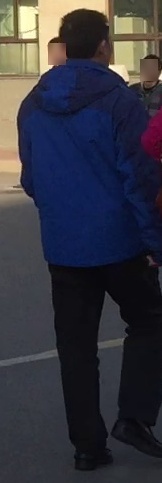}\\[-1pt]
    {\tiny\textbf{\textcolor{correct}{ID:~2055}}}
  \end{tabular} &
    \begin{tabular}{@{}c@{}}
    \adjincludegraphics[width=\imgw,height=\imgh,cfbox=correct 2pt 0pt]{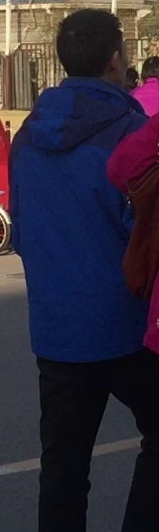}\\[-1pt]
    {\tiny\textbf{\textcolor{correct}{ID:~2055}}}
  \end{tabular} &
    \begin{tabular}{@{}c@{}}
    \adjincludegraphics[width=\imgw,height=\imgh,cfbox=correct 2pt 0pt]{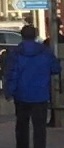}\\[-1pt]
    {\tiny\textbf{\textcolor{correct}{ID:~2055}}}
  \end{tabular} &
    \begin{tabular}{@{}c@{}}
    \adjincludegraphics[width=\imgw,height=\imgh,cfbox=correct 2pt 0pt]{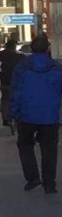}\\[-1pt]
    {\tiny\textbf{\textcolor{correct}{ID:~2055}}}
  \end{tabular} &
    \begin{tabular}{@{}c@{}}
    \adjincludegraphics[width=\imgw,height=\imgh,cfbox=correct 2pt 0pt]{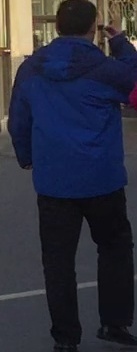}\\[-1pt]
    {\tiny\textbf{\textcolor{correct}{ID:~2055}}}
  \end{tabular} &
    \begin{tabular}{@{}c@{}}
    \adjincludegraphics[width=\imgw,height=\imgh,cfbox=correct 2pt 0pt]{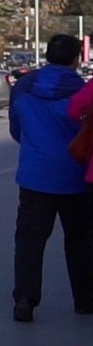}\\[-1pt]
    {\tiny\textbf{\textcolor{correct}{ID:~2055}}}
  \end{tabular} &
    \begin{tabular}{@{}c@{}}
    \adjincludegraphics[width=\imgw,height=\imgh,cfbox=correct 2pt 0pt]{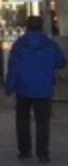}\\[-1pt]
    {\tiny\textbf{\textcolor{correct}{ID:~2055}}}
  \end{tabular} &
    \begin{tabular}{@{}c@{}}
    \adjincludegraphics[width=\imgw,height=\imgh,cfbox=correct 2pt 0pt]{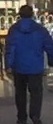}\\[-1pt]
    {\tiny\textbf{\textcolor{correct}{ID:~2055}}}
  \end{tabular} &
    \begin{tabular}{@{}c@{}}
    \adjincludegraphics[width=\imgw,height=\imgh,cfbox=correct 2pt 0pt]{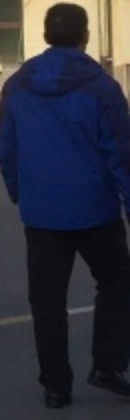}\\[-1pt]
    {\tiny\textbf{\textcolor{correct}{ID:~2055}}}
  \end{tabular} \\[2pt]
  \hline \\[0.1pt]

  \rotatebox{90}{\parbox{20mm}{\centering\footnotesize\textbf{No BRM}}} &
    \begin{tabular}{@{}c@{}}
    \adjincludegraphics[width=\imgw,height=\imgh,cfbox=querycol 2pt 0pt]{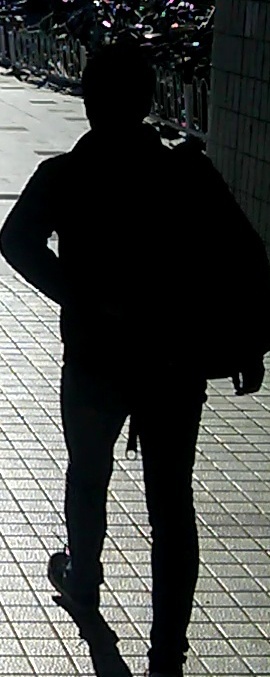}\\[-1pt]
    {\tiny\textbf{\textcolor{querycol}{ID:~2635}}}
  \end{tabular} &
    \begin{tabular}{@{}c@{}}
    \adjincludegraphics[width=\imgw,height=\imgh,cfbox=wrong 2pt 0pt]{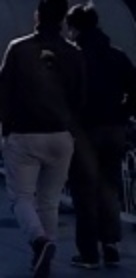}\\[-1pt]
    {\tiny\textbf{\textcolor{wrong}{ID:~1548}}}
  \end{tabular} &
    \begin{tabular}{@{}c@{}}
    \adjincludegraphics[width=\imgw,height=\imgh,cfbox=wrong 2pt 0pt]{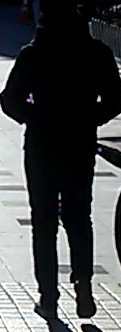}\\[-1pt]
    {\tiny\textbf{\textcolor{wrong}{ID:~2740}}}
  \end{tabular} &
    \begin{tabular}{@{}c@{}}
    \adjincludegraphics[width=\imgw,height=\imgh,cfbox=wrong 2pt 0pt]{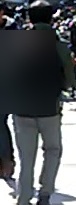}\\[-1pt]
    {\tiny\textbf{\textcolor{wrong}{ID:~1925}}}
  \end{tabular} &
    \begin{tabular}{@{}c@{}}
    \adjincludegraphics[width=\imgw,height=\imgh,cfbox=wrong 2pt 0pt]{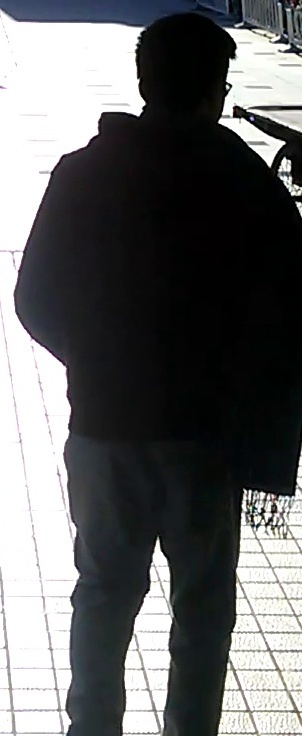}\\[-1pt]
    {\tiny\textbf{\textcolor{wrong}{ID:~2745}}}
  \end{tabular} &
    \begin{tabular}{@{}c@{}}
    \adjincludegraphics[width=\imgw,height=\imgh,cfbox=wrong 2pt 0pt]{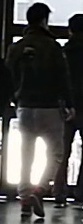}\\[-1pt]
    {\tiny\textbf{\textcolor{wrong}{ID:~1548}}}
  \end{tabular} &
    \begin{tabular}{@{}c@{}}
    \adjincludegraphics[width=\imgw,height=\imgh,cfbox=wrong 2pt 0pt]{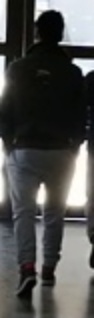}\\[-1pt]
    {\tiny\textbf{\textcolor{wrong}{ID:~1548}}}
  \end{tabular} &
    \begin{tabular}{@{}c@{}}
    \adjincludegraphics[width=\imgw,height=\imgh,cfbox=wrong 2pt 0pt]{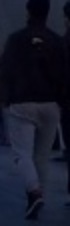}\\[-1pt]
    {\tiny\textbf{\textcolor{wrong}{ID:~1548}}}
  \end{tabular} &
    \begin{tabular}{@{}c@{}}
    \adjincludegraphics[width=\imgw,height=\imgh,cfbox=wrong 2pt 0pt]{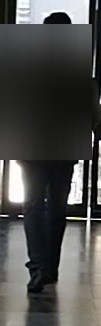}\\[-1pt]
    {\tiny\textbf{\textcolor{wrong}{ID:~1458}}}
  \end{tabular} &
    \begin{tabular}{@{}c@{}}
    \adjincludegraphics[width=\imgw,height=\imgh,cfbox=wrong 2pt 0pt]{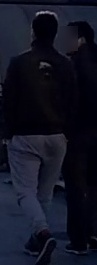}\\[-1pt]
    {\tiny\textbf{\textcolor{wrong}{ID:~1548}}}
  \end{tabular} &
    \begin{tabular}{@{}c@{}}
    \adjincludegraphics[width=\imgw,height=\imgh,cfbox=wrong 2pt 0pt]{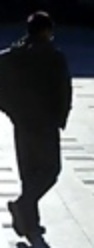}\\[-1pt]
    {\tiny\textbf{\textcolor{wrong}{ID:~2736}}}
  \end{tabular} \\[2pt]
  \rotatebox{90}{\parbox{20mm}{\centering\footnotesize\textbf{BRM}}} &
    \begin{tabular}{@{}c@{}}
    \adjincludegraphics[width=\imgw,height=\imgh,cfbox=querycol 2pt 0pt]{supp/images_msmt17/part3/q06_pid2635.jpg}\\[-1pt]
    {\tiny\textbf{\textcolor{querycol}{ID:~2635}}}
  \end{tabular} &
    \begin{tabular}{@{}c@{}}
    \adjincludegraphics[width=\imgw,height=\imgh,cfbox=correct 2pt 0pt]{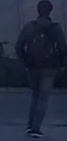}\\[-1pt]
    {\tiny\textbf{\textcolor{correct}{ID:~2635}}}
  \end{tabular} &
    \begin{tabular}{@{}c@{}}
    \adjincludegraphics[width=\imgw,height=\imgh,cfbox=correct 2pt 0pt]{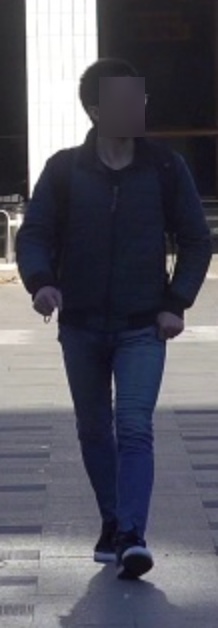}\\[-1pt]
    {\tiny\textbf{\textcolor{correct}{ID:~2635}}}
  \end{tabular} &
    \begin{tabular}{@{}c@{}}
    \adjincludegraphics[width=\imgw,height=\imgh,cfbox=wrong 2pt 0pt]{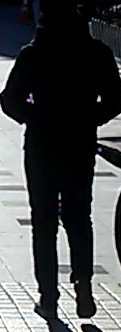}\\[-1pt]
    {\tiny\textbf{\textcolor{wrong}{ID:~2740}}}
  \end{tabular} &
    \begin{tabular}{@{}c@{}}
    \adjincludegraphics[width=\imgw,height=\imgh,cfbox=correct 2pt 0pt]{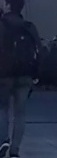}\\[-1pt]
    {\tiny\textbf{\textcolor{correct}{ID:~2635}}}
  \end{tabular} &
    \begin{tabular}{@{}c@{}}
    \adjincludegraphics[width=\imgw,height=\imgh,cfbox=correct 2pt 0pt]{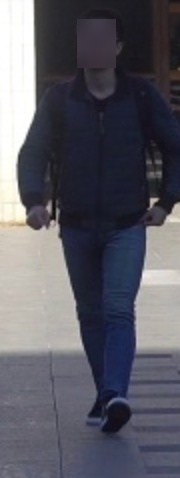}\\[-1pt]
    {\tiny\textbf{\textcolor{correct}{ID:~2635}}}
  \end{tabular} &
    \begin{tabular}{@{}c@{}}
    \adjincludegraphics[width=\imgw,height=\imgh,cfbox=correct 2pt 0pt]{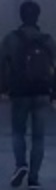}\\[-1pt]
    {\tiny\textbf{\textcolor{correct}{ID:~2635}}}
  \end{tabular} &
    \begin{tabular}{@{}c@{}}
    \adjincludegraphics[width=\imgw,height=\imgh,cfbox=correct 2pt 0pt]{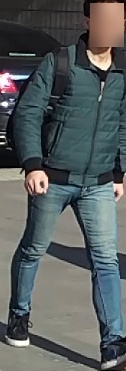}\\[-1pt]
    {\tiny\textbf{\textcolor{correct}{ID:~2635}}}
  \end{tabular} &
    \begin{tabular}{@{}c@{}}
    \adjincludegraphics[width=\imgw,height=\imgh,cfbox=correct 2pt 0pt]{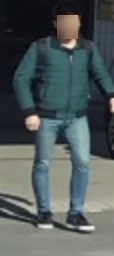}\\[-1pt]
    {\tiny\textbf{\textcolor{correct}{ID:~2635}}}
  \end{tabular} &
    \begin{tabular}{@{}c@{}}
    \adjincludegraphics[width=\imgw,height=\imgh,cfbox=wrong 2pt 0pt]{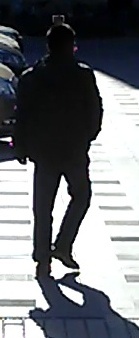}\\[-1pt]
    {\tiny\textbf{\textcolor{wrong}{ID:~2736}}}
  \end{tabular} &
    \begin{tabular}{@{}c@{}}
    \adjincludegraphics[width=\imgw,height=\imgh,cfbox=correct 2pt 0pt]{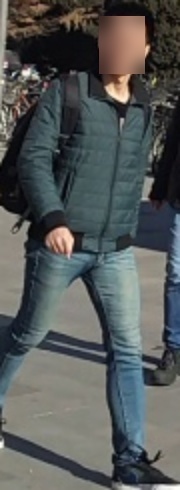}\\[-1pt]
    {\tiny\textbf{\textcolor{correct}{ID:~2635}}}
  \end{tabular} \\[2pt]
  \end{tabular}}

}
  \caption{%
    Top-10 retrieval results for 3 MSMT17 queries (PIDs 2046, 2055, 2635), comparing the baseline without BRM against the BRM-trained model.
    \textcolor{correct}{Green} = correct match;\;
    \textcolor{wrong}{red} = incorrect match;\;
    \textcolor{querycol}{blue} = query image.}
  \label{fig:retrieval-msmt17-02}
\end{figure*}